\documentclass[10pt,twocolumn,letterpaper]{article}

\usepackage{iccv}
\usepackage{times}
\usepackage{epsfig}
\usepackage{graphicx}
\usepackage{amsmath}
\usepackage{amssymb}

\usepackage{multirow}
\usepackage{xcolor}
\usepackage{array}
\usepackage{mathtools}
\usepackage{overpic}
\usepackage{stmaryrd}
\usepackage{booktabs}

\usepackage{caption}
\usepackage[symbol]{footmisc}

\usepackage[pagebackref,breaklinks,colorlinks]{hyperref}

\usepackage[capitalize]{cleveref}
\crefname{section}{Sec.}{Secs.}
\Crefname{section}{Section}{Sections}
\crefname{table}{Tab.}{Tabs.}
\Crefname{table}{Table}{Tables}
\crefname{figure}{Fig.}{Figs.}
\Crefname{figure}{Figure}{Figures}
\crefname{equation}{Eq.}{Eqs.}
\Crefname{equation}{Equation}{Equations}
\crefname{appendix}{Appx.}{Appxs.}
\Crefname{Appendix}{Appendix}{Appendices}

\iccvfinalcopy 

\def\iccvPaperID{9249} 

\ificcvfinal\fi

\newcommand{\ours}{2D3D-MATR}

\newcolumntype{P}[1]{>{\centering\arraybackslash}p{#1}}
\newlength{\wdth}

\newcommand{\ptitle}[1]{\noindent\textbf{#1}}

\begin{document}

\title{\ours{}: 2D-3D Matching Transformer for Detection-free Registration\\between Images and Point Clouds}


\author{
Minhao Li$^{1}$\footnotemark{}$\hspace{4pt}\hspace{8pt}$Zheng Qin$^{2,1}$\footnotemark[1]$\hspace{4pt}\hspace{8pt}$Zhirui Gao$^{1}\hspace{8pt}$Renjiao Yi$^{1}\hspace{8pt}$Chenyang Zhu$^{1}\hspace{8pt}$Yulan Guo$^{1,3}$\hspace{8pt}Kai Xu$^{1}$\footnotemark{}\\
$^{1}$National University of Defense Technology\\
$^{2}$Defense Innovation Institute, Academy of Military Sciences\hspace{10pt}$^{3}$Sun Yat-sen University\\
}

\twocolumn[{%
\renewcommand\twocolumn[1][]{#1}%
\maketitle
\begin{center}
\centering
\captionsetup{type=figure}
\vspace{-25pt}
\includegraphics[width=1\linewidth]{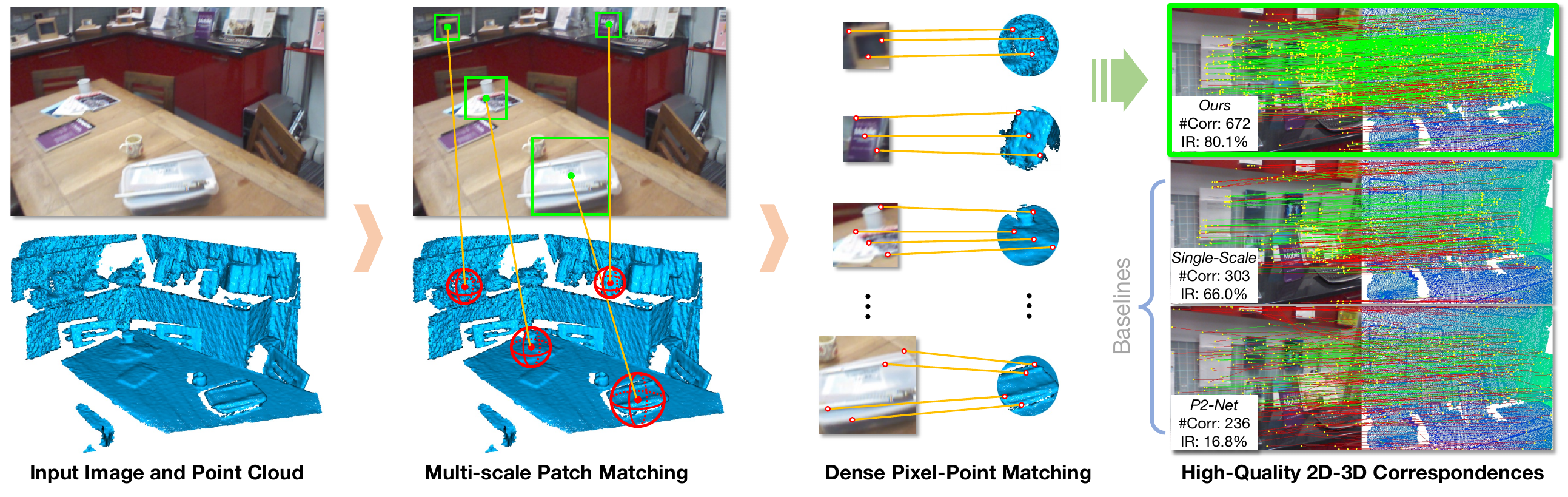}
\vspace{-20pt}
\captionof{figure}{
We propose \ours{}, a novel detection-free method for accurate inter-modality matching between images and point clouds.
Our method adopts a coarse-to-fine pipeline where it first computes coarse correspondences between downsampled image patches and point patches and then extends them to form dense pixel-point correspondences within the patch region.
A multi-scale sampling and matching scheme is devised to resolve the scale ambiguity in patch matching.
Compared to detection-based P2-Net (bottom-right) and single-scale patch matching (middle-right), \ours{} (top-right) extracts significantly more accurate and dense 2D-3D correspondences. The inliers are in {\color{green}green} and the outliers are in {\color{red}red}.
}
\label{fig:teaser}
\end{center}
}]

\footnotetext[1]{Equal contribution.}
\footnotetext[2]{Corresponding author: kevin.kai.xu@gmail.com.}


\begin{abstract}

The commonly adopted detect-then-match approach to registration finds difficulties in the cross-modality cases due to the incompatible keypoint detection and inconsistent feature description. We propose, \ours{}, a detection-free method for accurate and robust registration between images and point clouds.
Our method adopts a coarse-to-fine pipeline where it first computes coarse correspondences between downsampled patches of the input image and the point cloud and then extends them to form dense correspondences between pixels and points within the patch region.
The coarse-level patch matching is based on transformer which jointly learns global contextual constraints with self-attention and cross-modality correlations with cross-attention.
To resolve the scale ambiguity in patch matching, we construct a multi-scale pyramid for each image patch and learn to find for each point patch the best matching image patch at a proper resolution level.
Extensive experiments on two public benchmarks demonstrate that \ours{} outperforms the previous state-of-the-art P2-Net by around $20$ percentage points on inlier ratio and over $10$ points on registration recall. Our code and models are available at \url{https://github.com/minhaolee/2D3DMATR}.

\end{abstract}

\vspace{-15pt}


\section{Introduction}
\label{sec:introduction}

The inter-modality registration between images and point clouds finds applications in many computer vision tasks, e.g., 3D reconstruction, camera relocalization, SLAM and AR. It aims at estimating a rigid transformation that aligns a scene point cloud into the camera coordinates of an image capturing the same scene. The typical pipeline of 2D-3D registration is to first extract correspondences between pixels and points and then adopt robust pose estimators such as PnP-RANSAC~\cite{lepetit2009epnp,fischler1981random} to recover the alignment transformation. Therefore, the accuracy of the putative correspondences is the crux of a successful registration.

Following the intra-modality correspondence methods for stereo images~\cite{detone2018superpoint,luo2020aslfeat,sarlin2020superglue,dusmanu2019d2} or point clouds~\cite{gojcic2019perfect,choy2019fully,bai2020d3feat,huang2021predator}, 2D-3D matching methods~\cite{feng20192d3d,pham2020lcd,wang2021p2} usually adopt a \emph{detect-then-match} approach where 2D and 3D keypoints are first detected independently in the image and the point cloud, respectively, and then matched based on their associated descriptors. Such method, however, suffers from two difficulties. First, 2D and 3D keypoints are detected in different visual domains. While 2D keypoint detection is based on texture and color information, 3D detection is hinged on local geometry. This makes the detection of repeatable keypoints difficult. Second, 2D and 3D descriptors encode different visual information, which hampers extracting consistent descriptors for matching pixels and points. As a consequence, existing 2D-3D matching methods often lead to too low inlier ratio to be practically usable.

Recently, \emph{detection-free} approach has received increasing attention in both stereo matching~\cite{rocco2018neighbourhood,li2020dual,zhou2021patch2pix,sun2021loftr} and point cloud registration~\cite{yu2021cofinet,qin2022geometric}. Saving the step of keypoint detection, it achieves high-quality correspondence with a \emph{coarse-to-fine} pipeline: It first establishes coarse correspondences at the level of image or point patches and then refines them into fine-grained matching of pixels or points. This method has shown strong superiority over detection-based ones due to the exploitation of global contextual information at patch level. Such success, however, has not been attained for 2D-3D matching. This is because designing a coarse-level 2D-3D matching is non-trivial due to the scale ambiguity between image and point patches caused by perspective projection (see \cref{fig:teaser}). On the one hand, the receptive fields for extracting 2D and 3D features could be misaligned, resulting in inconsistency between 2D and 3D features. On the other hand, there could be many pixels or points finding no counterpart on other side due to occlusion, leading to considerable ambiguity for fine-level matching.

We propose \emph{\ours{}}, the first, to our knowledge, detection-free method for accurate and robust 2D-3D registration via addressing the challenges above.
Adapting the coarse-to-fine pipeline, our method first computes coarse correspondences between downsampled patches of the input image and the point cloud and then extends them to form dense correspondences between pixels and points within the patch regions.
To achieve accurate feature alignment betweem image and point patches, we design a coarse-level matching module based on transformer~\cite{vaswani2017attention} which jointly learns global contextual constraints with self-attention and cross-modality correlations with cross-attention.

Our key insight is that the feature misalignment between 2D and 3D due to projection can be resolved by \emph{image-space multi-scale sampling and matching}, assuming that the area of local patches is small and the projection distortions is negligible. We construct a multi-scale pyramid for each image patch. During training, we find for each point patch the best matching image patch at a proper resolution level through computing the bilateral overlap between them in the image space. During test, our model can automatically infer 2D-3D patch correspondences at a proper scale and produces dense correspondence in a high inlier ratio.
Extensive experiments on the RGB-D Scenes V2~\cite{lai2014unsupervised} and 7-Scenes~\cite{glocker2013real} benchmarks demonstrate the efficacy of our method. In particular, \ours{} outperforms the previous state-of-the-art P2-Net~\cite{wang2021p2} by at least $20$ percentage points on inlier ratio and over $10$ points on registration recall on the two benchmarks.
Our contributions include:
\begin{itemize}
	\vspace{-8pt}\item The first detection-free coarse-to-fine matching network for 2D-3D registration which first establishes coarse correspondences of patch level and then refines them into dense correspondences of pixel/point level.
	\vspace{-8pt}\item A transformer-based coarse matching module learning well-aligned 2D and 3D features with both global contextual constraints and cross-modality correlations.
	\vspace{-8pt}\item A multi-scale 2D-3D matching scheme that resolves 2D-3D feature misalignment through learning image-space multi-scale features and feature-scale selection.
\end{itemize}


\begin{figure*}[t]
\centering
\includegraphics[width=\linewidth]{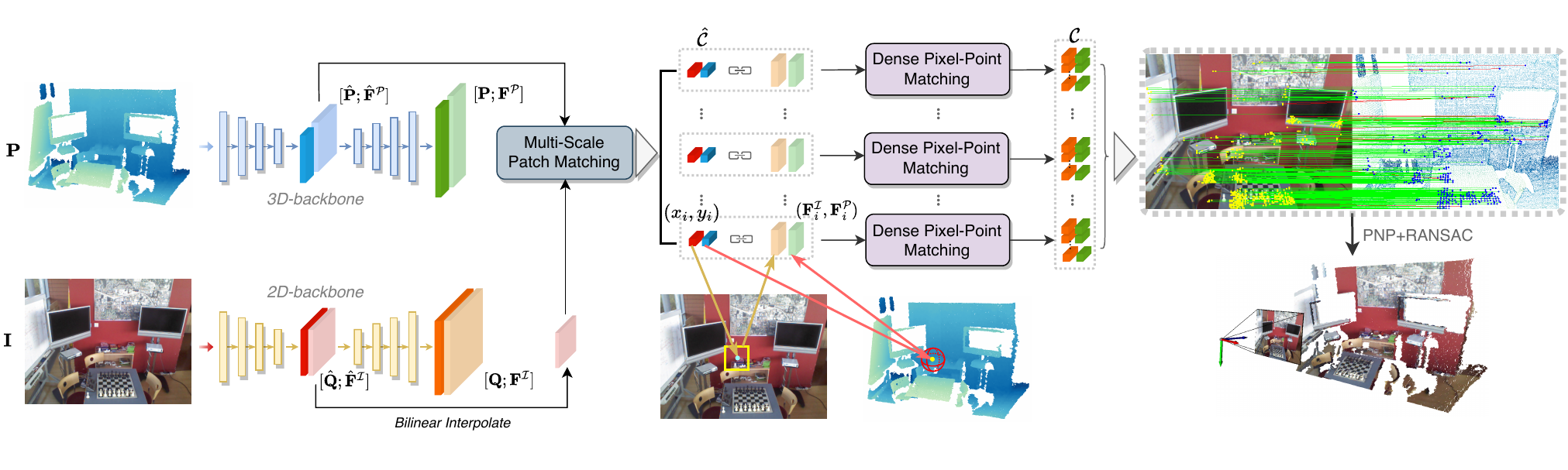}
\vspace{-25pt}
\caption{
Overall pipeline of \ours{}.
We first progressively downsample the input image $\mathbf{I}$ and the point cloud $\mathbf{P}$ and learn multi-scale 2D and 3D features. The 2D and 3D features $\hat{\mathbf{F}}^{\mathcal{I}}$ and $\hat{\mathbf{F}}^{\mathcal{P}}$ at the coarsest stage are used to extract coarse correspondences between the local patches of the image and the point cloud. A multi-scale patch matching module is devised to learn global contextual constraints and cross-modality correlations. Next, the patch correspondences are extended to dense pixel-point correspondences based on the high-resolution features $\mathbf{F}^{\mathcal{I}}$ and $\mathbf{F}^{\mathcal{P}}$. Finally, PnP-RANSAC is adopted to estimate the alignment transformation.
}
\label{fig:pipeline}
\vspace{-15pt}
\end{figure*}


\section{Related Work}
\label{sec:related work}

\ptitle{Stereo image registration.}
Traditional stereo image registration methods usually adopt a \emph{detect-then-match} pipeline to extract correspondences. A set of sparse keypoints are first detected and described with hand-crafted~\cite{lowe1999object,rublee2011orb} or learning-based descriptors~\cite{revaud2019r2d2,dusmanu2019d2,detone2018superpoint,sarlin2020superglue,luo2020aslfeat} from both sides, which are then matched based on feature similarity.
Keypoints detection is ill-posed and detection-free methods~\cite{rocco2018neighbourhood,rocco2020efficient,lee2021patchmatch} propose to bypass the keypoint detection step by computing a correlation matrix between all pairs of features. However, the all-pair correlation matrix requires huge computation, making the putative correspondences relatively coarse-grained. For this reason, \cite{li2020dual,zhou2021patch2pix,sun2021loftr} further propose to adopt a coarse-to-fine matching framework, which achieves accurate and efficient image matching.

\ptitle{Point cloud registration.}
Similar progress as in image registration has also been witnessed in point cloud registration.
Early works leverage hand-crafted descriptors such as PPF~\cite{drost2010model} and FPFH~\cite{rusu2009fast} for keypoint detection. And recent learning-based descriptors~\cite{deng2018ppfnet,deng2018ppf,choy2019fully,gojcic2019perfect,bai2020d3feat,huang2021predator} achieve more robust and accurate matching results.
To bypass the keypoint detection, CoFiNet~\cite{yu2021cofinet} introduces the coarse-to-fine strategy to the matching of point clouds. And GeoTransformer~\cite{qin2022geometric} further designs a transformation-invariant geometric structure embedding and achieves RANSAC-free point cloud registration.
Moreover, there are also methods focusing on removing outlier correspondences~\cite{bai2021pointdsc,choy2020deep,lee2021deep}, which act as an effective alternative of traditional robust estimators such as RANSAC~\cite{fischler1981random}.

\ptitle{Inter-modality registration.}
Compared to intra-modality matching problems, inter-modality matching between images and point clouds is more difficult.
Based on how the correspondences are established, previous works can be classified into two categories. The first class focuses on visual localization in a known scene. The main idea of them is to predict the 3D coordinates of each image pixel with decision trees~\cite{shotton2013scene,meng2017backtracking,meng2018exploiting,brachmann2016uncertainty,valentin2015exploiting} or neural networks~\cite{brachmann2017dsac,brachmann2018learning,brachmann2019neural,li2020hierarchical,li2018full,massiceti2017random,yang2019sanet}.
However, this class of methods lack generality to novel scenes.
The second class follows the traditional detect-then-match pipeline~\cite{feng20192d3d,pham2020lcd,wang2021p2}, where keypoints are first detected from each modality and then matched with the associated descriptors.
Compared to the first class, this class of methods have better generality theoretically.
However, detecting repeatable inter-modality keypoints is much more difficult and unstable as keypoints are defined and described in different visual domains. For this reason, existing methods still suffer from low inlier ratio.
In this work, we propose \ours{} to address these issues with two specific designs, \ie, coarse-to-fine matching and transformer-based multi-scale patch matching.


\section{Method}

\subsection{Overview}

Given a image $\mathbf{I} \in \mathbb{R}^{H \times W \times 3}$ and a point cloud $\mathbf{P} \in \mathbb{R}^{N \times 3}$ of a scene, the goal of 2D-3D registration is to recover the alignment transformation $\mathbf{T}$ between them, which is composed of a 3D rotation $\mathbf{R} \in \mathcal{SO}(3)$ and a 3D translation $\mathbf{t} \in \mathbf{R}^3$.
A traditional 2D-3D registration pipeline first extracts correspondences $\mathcal{C} = \{ (\mathbf{x}_i, \mathbf{y}_i) \mid \mathbf{x}_i \in \mathbb{R}^3, \mathbf{y}_i \in \mathbb{R}^2 \}$ between 3D points and 2D pixels, and then estimates the transformation by minimizing the 2D projection error:
\begin{equation}
\min_{\mathbf{R}, \mathbf{t}} \sum_{(\textbf{x}_i, \textbf{y}_i) \in \mathcal{C}} \lVert \mathcal{K}(\mathbf{R}\mathbf{x}_i + \mathbf{t}, \mathbf{K}) - \mathbf{y}_i \rVert^2,
\end{equation}
where $\mathbf{K}$ is the intrinsic matrix of the camera and $\mathcal{K}$ is the project function from 3D space to image plane.
This problem can be effectively solved by PnP-RANSAC algorithm. However, the solution can be erroneous due to inaccurate correspondences.

In this work, we present a method to hierarchically extract inter-modality correspondences.
We first adopt two respective backbones to learn features for the image and point cloud (\cref{sec:feature-extraction}).
Next, we extract a set of coarse correspondences between the downsampled patches of the image and the point cloud (\cref{sec:patch-matching}).
At last, the patch correspondences are the refined to dense pixel-point correspondences on the fine level (\cref{sec:dense-matching}).
\cref{fig:pipeline} illustrates the overall pipeline of our method.


\begin{figure*}[t]
\centering
\includegraphics[width=\linewidth]{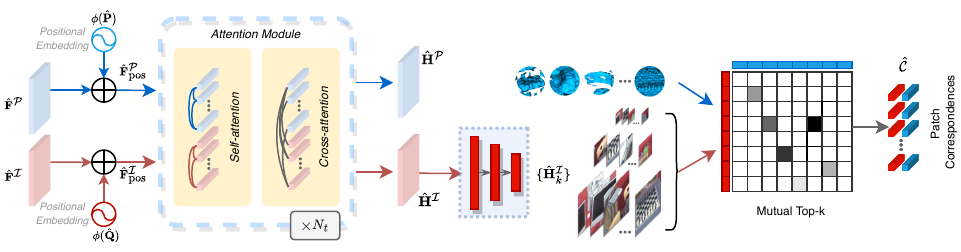}
\vspace{-20pt}
\caption{
Multi-scale patch matching.
Given the coarse 2D and 3D features, we first learn global contextual constraints with self-attention and cross-modality correlations with cross-attention. Then we adopt an image-space multi-scale sampling and matching strategy to extract patch correspondences which are better aligned in the image plane.
}
\label{fig:patch-matching}
\vspace{-10pt}
\end{figure*}

\subsection{Feature Extraction}
\label{sec:feature-extraction}

\ptitle{Backbones.}
Given a pair of image and point cloud, two modality-specific encoder-decoder backbone networks are adopted for hierarchical feature extraction.
For the image, we use a ResNet~\cite{he2016deep} with FPN~\cite{lin2017feature} to generate multi-scale image features.
The downsampled 2D features $\hat{\mathbf{F}}^{\mathcal{I}} \in \mathbb{R}^{\hat{H} \times \hat{W} \times \hat{C}}$ at the smallest resolution and $\mathbf{F}^{\mathcal{I}} \in \mathbb{R}^{H \times W \times C}$ at the original resolution are used for matching in coarse and fine levels.
For simplicity, we denote the pixel coordinate matrices for $\hat{\mathbf{F}}^{\mathcal{I}}$ and $\mathbf{F}^{\mathcal{I}}$ as $\hat{\mathbf{Q}} \in \mathbb{R}^{\hat{H} \times \hat{W} \times 2}$ and $\mathbf{Q} \in \mathbb{R}^{H \times W \times 2}$, respectively.
For the point cloud, we adopt KPFCNN~\cite{thomas2019kpconv} to learn 3D features following~\cite{bai2020d3feat,huang2021predator,yu2021cofinet,qin2022geometric}.
Unlike images which have fixed resolutions, point clouds usually have inconsistent sizes and KPFCNN dynamically downsamples them via grid downsampling.
We use the points $\hat{\mathbf{P}} \in \mathbf{R}^{\hat{N} \times 3}$ corresponding to the coarsest level and their associated features $\hat{\mathbf{F}}^{\mathcal{P}} \in \mathbb{R}^{\hat{N} \times \hat{C}}$ for coarse-level matching, while fine-level matching is conducted on the input point cloud $\mathbf{P}$ and the associated features $\mathbf{F}^{\mathcal{P}} \in \mathbb{R}^{N \times C}$.

\ptitle{Patch construction.}
To extract patch correspondences on the coarse level, we need first associate each downsampled pixel (point) with an image (point) patch.
For the image, we evenly divide $\mathbf{I}$ into $\hat{H} \times \hat{W}$ patches and each pixel in $\hat{\mathbf{F}}$ corresponds to an \emph{image patch} of $\frac{H}{\hat{H}} \times \frac{W}{\hat{W}}$ pixels. 
For the point cloud, we use point-to-node partition~\cite{li2018so} following~\cite{yu2021cofinet,qin2022geometric}, which assigns each point in $\mathbf{P}$ to its nearest point in $\hat{\mathbf{P}}$ to compose the \emph{point patches}.

\subsection{Multi-scale Patch Matching}
\label{sec:patch-matching}

\ptitle{Attention-based feature refinement.}
Given the downsampled image $(\hat{\mathbf{Q}}, \hat{\mathbf{F}}^{\mathcal{I}})$ and point cloud $(\hat{\mathbf{P}}, \hat{\mathbf{F}}^{\mathcal{P}})$, our goal in the coarse level is to extract patch correspondences that overlap with each other.
However, inter-modality matching between 2D and 3D is non-trivial. On the one hand, 2D and 3D features are learned from different domains, leading to severe inconsistency between them.
This problem is more serious in patch matching than point matching as patch features are learned from a large context, which aggravates the feature misalignment.
Second, as noted in~\cite{sun2021loftr,yu2021cofinet,qin2022geometric}, coarse-level matching relaxes the matching criterion from the strict 3D distance into a much looser local texture-geometry similarity. This effectively eases the matching difficulty but requires more global context. For this reason, we devise a transformer-based~\cite{vaswani2017attention} feature refinement module to learn global contextual constraints and cross-modality correlations.

Before feeding into transformer, we first augment the 2D and 3D features with their positional information:
\begin{equation}
\hat{\mathbf{F}}^{\mathcal{I}}_{\text{pos}} = \hat{\mathbf{F}}^{\mathcal{I}} + \phi(\hat{\mathbf{Q}}),\hspace{10pt}\hat{\mathbf{F}}^{\mathcal{P}}_{\text{pos}} = \hat{\mathbf{F}}^{\mathcal{P}} + \phi(\hat{\mathbf{P}}),
\end{equation}
and $\phi(\cdot)$ is the Fourier embedding function~\cite{mildenhall2020nerf}:
\begin{equation}
\phi(x) {=} \bigl[x, \sin(2^0 x), \cos(2^0 x), ..., \sin(2^{L{-}1} x), \cos(2^{L{-}1} x)\bigr],
\end{equation}
where $L$ is the length of the embedding.
We then flatten the first two spatial dimensions of the 2D features for simplicity and use $\hat{\mathbf{F}}^{\mathcal{I}}_{\text{pos}}$, $\hat{\mathbf{F}}^{\mathcal{P}}_{\text{pos}}$ for future computation.

Afterwards, we leverage transformer to further refine the features in two modalities. Given anchor features $\mathbf{F}^{\mathcal{A}} \in \mathbb{R}^{\lvert \mathcal{A} \rvert \times d}$ and memory features $\mathbf{F}^{\mathcal{M}} \in \mathbb{R}^{\lvert \mathcal{M} \rvert \times d}$, transformer models the pairwise correlations between them with attention mechanism to generate more discriminative features. Specifically, the two set of features are first projected as:
\begin{equation}
\mathbf{Q} = \mathbf{F}^{\mathcal{A}} \mathbf{W}^Q,\hspace{5pt}\mathbf{K} = \mathbf{F}^{\mathcal{M}} \mathbf{W}^K,\hspace{5pt}\mathbf{V} = \mathbf{F}^{\mathcal{M}} \mathbf{W}^V,
\end{equation}
where $\mathbf{W}^Q, \mathbf{W}^K, \mathbf{W}^Q \in \mathbf{R}^{d \times d}$ are the projection weights for query, key and value. The attention features for the anchor set are then computed as:
\begin{equation}
\mathtt{Attention}(\mathbf{F}^{\mathcal{A}}, \mathbf{F}^{\mathcal{M}}) = \mathtt{Softmax}(\frac{\mathbf{Q}\mathbf{K}^T}{d^{0.5}})\mathbf{V}.
\end{equation}
And the attention features are further projected with a shallow MLP as the final output features.

We iteratively apply self-attention and cross-attention to refine the 2D and 3D features as shown in \cref{fig:patch-matching}. In self-attention, we use the features from the same modality as both the anchor features and memory features for attention computation to encode intra-modality global contextual constraints. In cross-attention, we use the features from one modality as the anchor features and the other modality as the memory features to learn cross-modality correlations. By this means, we can obtain refined 2D and 3D features which are more discriminative and better aligned. The resultant features are denoted as $\hat{\mathbf{H}}^{\mathcal{I}} \in \mathbb{R}^{\hat{H} \times \hat{W} \times \hat{C}}$ and $\hat{\mathbf{H}}^{\mathcal{P}} \in \mathbb{R}^{\lvert \hat{N} \rvert \times \hat{C}}$ in 2D and 3D modalites, respectively.

\begin{figure}[t]
\centering
\includegraphics[width=0.95\linewidth]{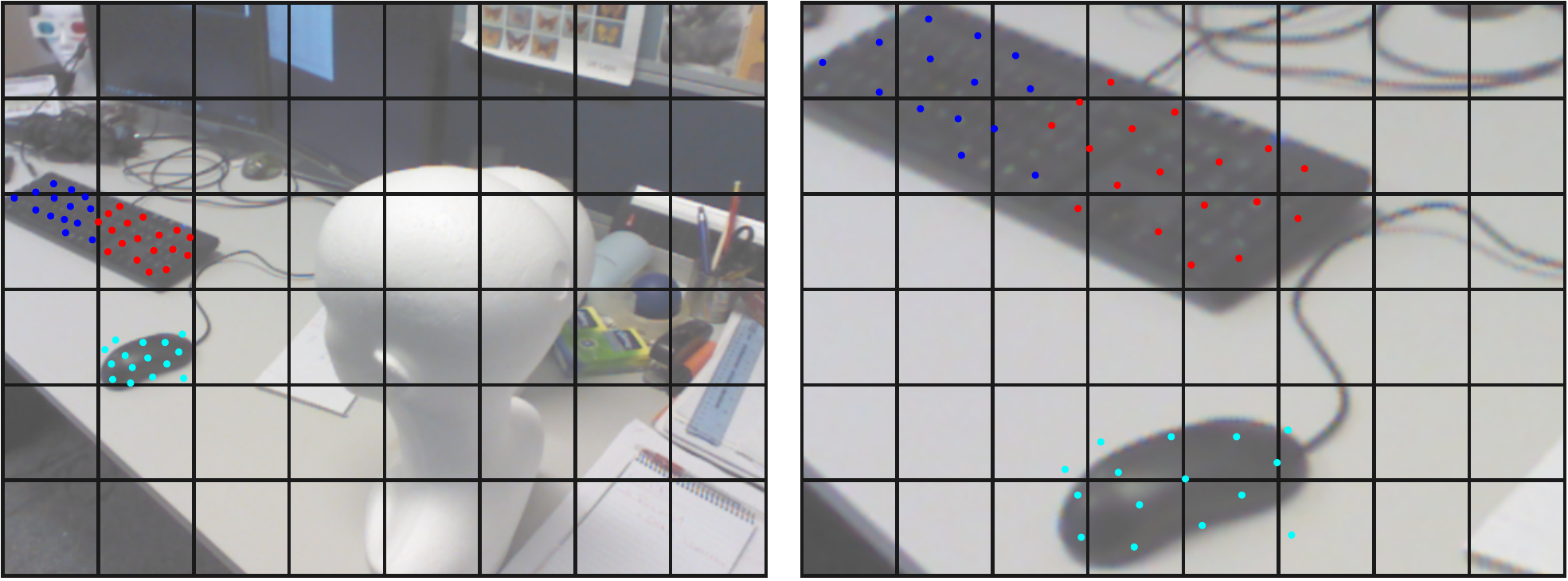}
\vspace{-5pt}
\caption{
Scale misalignment between image patches and point patches due to perspective projection. \textbf{Left}: when the camera is far from the scene, the 3D patches on the keyboard are properly aligned with the 2D patches, and the 3D patch around the mouse is even slightly smaller than the matched 2D patch. \textbf{Right}: when camera move towards the scene, the 3D patches cover several 2D patches, leading to severe matching ambiguity.
}
\label{fig:patch-misalignment}
\vspace{-15pt}
\end{figure}

\ptitle{Multi-scale matching.}
Due to the effect of perspective projection, the objects in images have the \emph{scale ambiguity} problem, \ie, an object looks larger if it lies close to the camera and smaller if far from the camera. However, the scale of an object in the point cloud remains unchanged and is agnostic to camera motion. As a result, the 2D and 3D patches could be seriously misaligned: a 3D patch could cover many 2D patches when the camera moves close, and vice versa. \cref{fig:patch-misalignment} illustrates the misalignment between 2D and 3D patches. This causes significant ambiguous objective during training: considering two nearby point patches with different physical properties, they could be supervised to have similar features if covered by the same image patch.
This is unexpected as it aggravates the feature misalignment and harms the distinctiveness of the features.


For this reason, we devise an image-space multi-scale sampling and matching strategy to alleviate the scale ambiguity between 2D and 3D patches. Technically, we first divide $\mathbf{I}$ into $\hat{H}_0 \times \hat{W}_0$ patches and then build a $K$-level patch pyramid for each image patch. At each pyramid level, the patch size is halved to generate a more fine-grained patch partition. 
The features of the patch pyramid is obtained by a lightweight $K$-stage CNN.
We first downsample $\hat{\mathbf{H}}^{\mathcal{I}}$ to fit the finest patch pyramid level. The 2D features are then downsampled by a factor of $2$ at each stage to match the resolutions of each patch pyramid level. For simplicity, the 2D patch features at the $k^{\text{th}}$ level are denoted as $\hat{\mathbf{H}}^{\mathcal{I}}_{k}$.
At last, the multi-scale 2D patch features $\{\hat{\mathbf{H}}^{\mathcal{I}}_{k}\}$ and the 3D patch features $\hat{\mathbf{H}}^{\mathcal{P}}$ are normalized onto a unit hypersphere as the final features.

\begin{figure}[t]
\centering
\includegraphics[width=\linewidth]{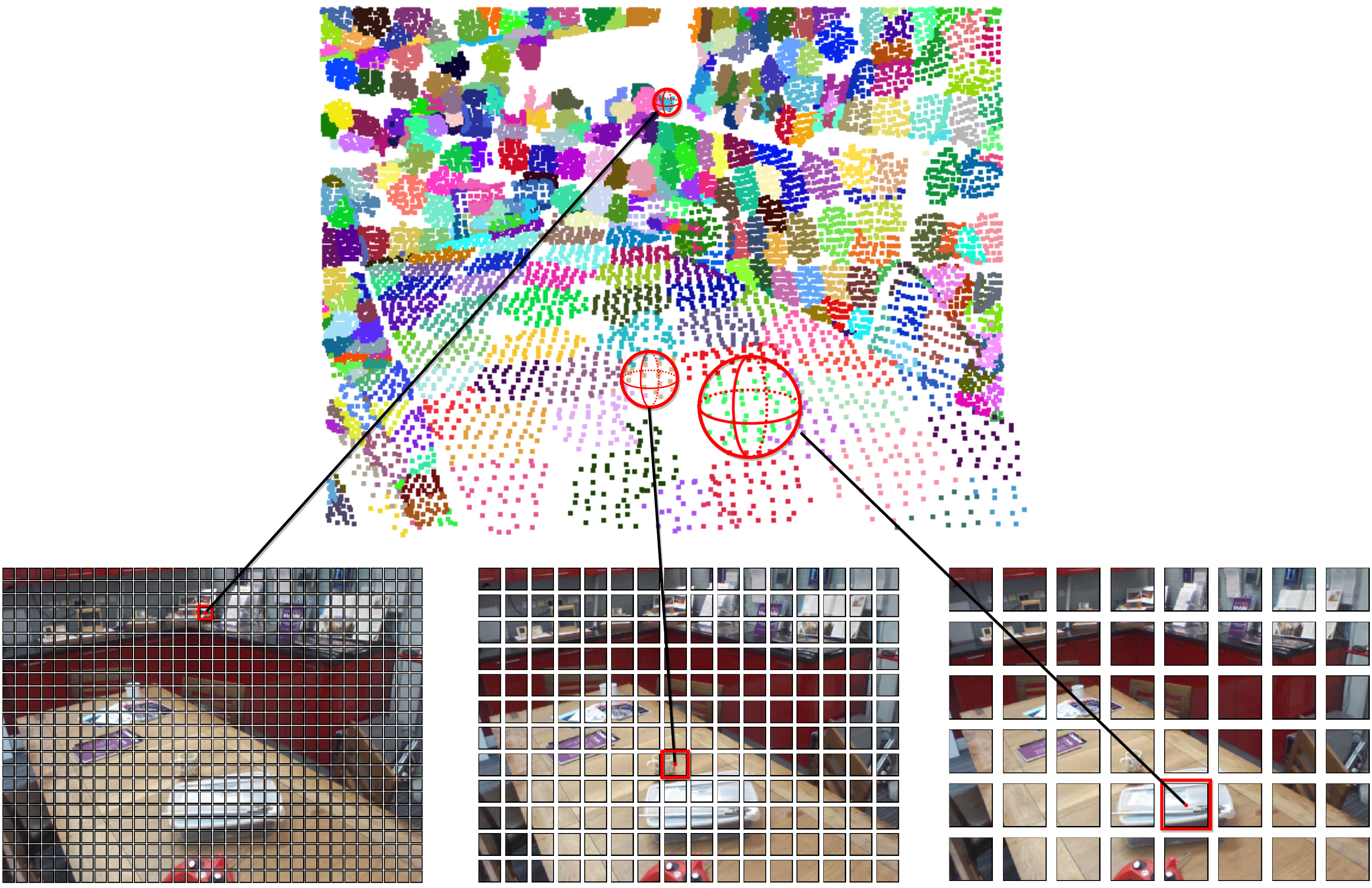}
\vspace{-15pt}
\caption{
Multi-scale patch matching based on image-space patch pyramid with $3$ levels.
One matched patch pair is shown in each pyramid level.
The 3D patches far from the camera are matched to small 2D patches in a later level, while the close ones are matched to large 2D patches in a early level.
}
\label{fig:multi-scale-matching}
\vspace{-10pt}
\end{figure}

By leveraging the multi-scale matching strategy, for each 3D patch, we find the 2D patch that coincides the best with it on the image plane during training:
the 3D patches far from the camera prefer small 2D patches in a later level, while the close ones are more likely to match with large 2D patches in a early level.
\cref{fig:multi-scale-matching} illustrates our multi-scale matching strategy, where our method provides 3D patches with better aligned 2D patches.
This can effectively alleviate the matching ambiguity and reduce the difficulty in learning consistent 2D and 3D features.
During inference, the putative patch correspondences $\hat{\mathcal{C}}$ are extracted with mutual top-$k$ selection~\cite{qin2022geometric}:
\begin{equation}
\begin{aligned}
(x_i, y_i)\text{ is matched} \Leftrightarrow & \bigl(\hat{\mathbf{h}}^{\mathcal{I}}_{*}(x_i)\text{ is }k\text{NN of }\hat{\mathbf{h}}^{\mathcal{P}}(y_i)\bigr)\land \\
&\bigl(\hat{\mathbf{h}}^{\mathcal{P}}(y_i)\text{ is }k\text{NN of }\hat{\mathbf{h}}^{\mathcal{I}}_{*}(x_i)\bigr)
\end{aligned}
\end{equation}


\subsection{Dense Pixel-Point Matching}
\label{sec:dense-matching}

After obtaining the patch correspondences, we further refine them to dense pixel-point correspondences. For each $(x_i, y_i) \in \hat{\mathcal{C}}$, we collect the fine-level 2D and 3D features of its local pixels and points from $\mathbf{F}^{\mathcal{I}}$ and $\mathbf{F}^{\mathcal{P}}$, denoted as $\mathbf{F}^{\mathcal{I}}_i$ and $\mathbf{F}^{\mathcal{P}}_i$. For computational efficiency, we uniformly sample $1/4$ of the pixels in each 2D patch. Following~\cref{sec:patch-matching}, we normalize $\mathbf{F}^{\mathcal{I}}_i$ and $\mathbf{F}^{\mathcal{P}}_i$ to unit-length vectors and match the pixels and points with mutual top-$k$ selection as the \emph{local dense correspondences} of $(x_i, y_i)$. We do not adopt a specific matching layer such as Sinkhorn~\cite{sarlin2020superglue,yu2021cofinet,qin2022geometric} here as the 2D patches in large scales could have enormous pixels (\eg, $1600$ pixels in our experiments), which causes unacceptable computational cost. On the contrary, mutual top-$k$ selection is very efficient and still achieves promising performance. At last, we gather the local correspondences of all $(x_i, y_i)$ from $\hat{\mathcal{C}}$ as the final dense pixel-point correspondences. Note that as the 2D patches from different scales can overlap with each other, we explicitly remove the repeated correspondences from the final correspondences.

\subsection{Loss Functions}

Our model is trained in a metric learning fashion. On the coarse level, we adopt a scaled circle loss~\cite{sun2020circle,qin2022geometric} to supervise the patch features. On the fine level, another standard circle loss~\cite{sun2020circle} is used to supervise the dense pixel and point features. The overall loss is then computed as $\mathcal{L}_{\text{all}} = \mathcal{L}_{\text{coarse}} + \lambda \mathcal{L}_{\text{fine}}$, where $\lambda = 1$ is a balance factor.

Compared to contrastive loss~\cite{chopra2005learning} and triplet loss~\cite{hoffer2014deep}, circle loss~\cite{sun2020circle} has a circular decision boundary which facilitates convergence.
Given an anchor descriptor $\mathbf{d}_{i}$, the descriptors of its positive and negative pairs are denoted as $\mathcal{D}^{\mathcal{P}}_{i}$ and $\mathcal{D}^{\mathcal{N}}_{i}$. The general circle loss on $\mathbf{d}_i$ is computed as:
\begin{equation}
\mathcal{L}_{i} {=} \frac{1}{\gamma} \log\bigl[ 1 {+} \sum_{\mathclap{\mathbf{d}_{j} \in \mathcal{D}^{\mathcal{P}}_{i}}} {e}^{ \beta^{i, j}_{p} (d^{j}_{i} - \Delta_{p})} {\cdot} \sum_{\mathclap{\mathbf{d}_{k} \in \mathcal{D}^{\mathcal{N}}_{i}}} {e}^{ \beta^{i, k}_{n} (\Delta_{n} - d^{k}_{i})} \bigr],
\end{equation}
where $d^{j}_{i}$ is the $\ell_{2}$ feature distance, $\beta^{i, j}_{p} = \gamma \lambda^{i, j}_{p} (d^{j}_{i} - \Delta_{p})$ and $\beta^{i, k}_{n} = \gamma \lambda^{i, k}_{n} (\Delta_{n} - d^{k}_{i})$ are the individual weights for the positive and negative pairs, where $\lambda^{i, j}_{p}$ and $\lambda^{i, k}_{n}$ are the scaling factors for the positive and negative pairs. 

On the coarse level, we generate the ground truth based on the bilateral overlap. A patch pair is regarded as positive if the 2D and 3D overlap ratios between them are both at least $30\%$, and as negative if both the overlap ratios are below $20\%$. Please refer to~\cref{sec:implementation} for more details.
The overlap ratio between the 2D and 3D patches are used as $\lambda_{p}$, and $\lambda_{n}$ is set to $1$. On the fine level, a pixel-point pair is positive with the 3D distance is below $3.75$cm and the 2D distance is below $8$ pixels, while being negative with a 3D distance above $10$cm or a 2D distance above $12$ pixels. The scaling factors are all $1$. We ignore all other pairs on both levels during training as the safe region. The margins are set to $\Delta_{p} = 0.1$ and $\Delta_{n} = 1.4$ following~\cite{huang2021predator,qin2022geometric}.


\section{Experiments} 
\label{sec:experiments}

As there is no public 2D-3D registration benchmark, we build two challenging benchmarks based on RGB-D Scenes V2~\cite{lai2014unsupervised} (\cref{sec:exp-rgbd}) and 7Scenes~\cite{glocker2013real} (\cref{sec:exp-7scenes}) datasets, and evaluate the efficacy of \ours{} on them. Extensive ablation studies are provided to study the influence of different design choices (\cref{sec:exp-ablation}).

\subsection{Implementation Details}
\label{sec:implementation}

\ptitle{Network architecture.}
We adopt a $4$-stage ResNet~\cite{he2016deep} with FPN as the image backbone network, where the output channels of each stage are $\{128, 128, 256, 512\}$. The resolution of the input images is $480 \times 640$ and is downsampled to $60 \times 80$ in the coarsest level. For the 3D backbone, we use a $4$-stage KPFCNN~\cite{thomas2019kpconv} where the output channels of each stage are $\{ 128, 256, 512, 1024 \}$. The point clouds are voxelized with an initial voxel size of $2.5$cm which is doubled at each stage.
In the coarse level, we resize the 2D features to $24 \times 32$ before feeding them to the transformer for computational efficiency.
All the transformer layers have $256$ features channels with $4$ attention heads and ReLU activation.
In the patch pyramid, we use $H_0 = 6$ and $W_0 = 8$ in the coarsest level and build $K=3$ pyramid levels, \ie, $\{ 6 \times 8, 12 \times 16, 24 \times 32 \}$. In the fine level, we project both the 2D and 3D features to $128$-d for feature matching.

\ptitle{Metrics.}
We mainly evaluate the models with $3$ metrics:
(1) \emph{Inlier Ratio} (IR), the ratio of pixel-point matches whose 3D distance is below a certain threshold (\ie, $5$cm) over all putative matches.
(2) \emph{Feature Matching Recall} (FMR), the ratio of image-point-cloud pairs whose inlier ratio is above a certain threshold (\ie, $10\%$).
(3) \emph{Registration Recall} (RR), the ratio of image-point-cloud pairs whose $\mathtt{RMSE}$ is below a certain threshold (\ie, $10$cm).

\ptitle{Baselines.}
We mainly compare to $3$ keypoint detection-based baseline methods:
(1) FCGF2D3D, a 2D-3D implementation of FCGF~\cite{choy2019fully} which samples random keypoints from the image and the point cloud.
(2) Predator2D3D, a 2D-3D implementation of Predator~\cite{huang2021predator} which leverages a graph network to learn the saliency of each pixel (point) for sampling keypoints.
(3) P2-Net~\cite{wang2021p2}, a 2D-3D correspondence method which detects locally salient pixels (points) in the feature space.
Note that albeit successful in point cloud registration, we find that Predator-2D3D fails to predict reliable saliency scores in 2D-3D scenarios. To this end, we ignore the saliency scores in Predator-2D3D and randomly sample keypoints according to the predicted overlap scores.
For fair comparison, we use the same backbones for all the methods.
Please refer to Appx.~A for more details.

\subsection{Evaluations on RGB-D Scenes V2}
\label{sec:exp-rgbd}


\begin{table}[!t]
\scriptsize
\setlength{\tabcolsep}{4pt}
\centering
\begin{tabular}{l|ccccc}
\toprule
Model & Scene-11 & Scene-12 & Scene-13 & Scene-14 & Mean \\
\midrule
Mean depth (m) & 1.74 & 1.66 & 1.18 & 1.39 & 1.49 \\
\midrule
\multicolumn{6}{c}{\emph{Inlier Ratio} $\uparrow$} \\
\midrule
FCGF-2D3D~\cite{choy2019fully} & 6.8 & 8.5 & 11.8 & 5.4 & 8.1 \\
P2-Net~\cite{wang2021p2} & 9.7 & 12.8 & 17.0 & \underline{9.3} & 12.2 \\
Predator-2D3D~\cite{huang2021predator} & \underline{17.7} & \underline{19.4} & \underline{17.2} & 8.4 & \underline{15.7} \\
\ours{} (\emph{ours}) & \textbf{32.8} & \textbf{34.4} & \textbf{39.2} & \textbf{23.3} & \textbf{32.4} \\
\midrule
\multicolumn{6}{c}{\emph{Feature Matching Recall} $\uparrow$} \\
\midrule
FCGF-2D3D~\cite{choy2019fully} & 11.1 & 30.4 & 51.5 & 15.5 & 27.1 \\
P2-Net~\cite{wang2021p2} & 48.6 & 65.7 & \underline{82.5} & \underline{41.6} & 59.6 \\
Predator-2D3D~\cite{huang2021predator} & \underline{86.1} & \underline{89.2} & 63.9 & 24.3 & \underline{65.9} \\
\ours{} (\emph{ours}) & \textbf{98.6} & \textbf{98.0} & \textbf{88.7} & \textbf{77.9} & \textbf{90.8} \\
\midrule
\multicolumn{6}{c}{\emph{Registration Recall} $\uparrow$} \\
\midrule
FCGF-2D3D~\cite{choy2019fully} & 26.4 & 41.2 & 37.1 & 16.8 & 30.4 \\
P2-Net~\cite{wang2021p2} & 40.3 & 40.2 & \underline{41.2} & \underline{31.9} & \underline{38.4} \\
Predator-2D3D~\cite{huang2021predator} & \underline{44.4} & \underline{41.2} & 21.6 & 13.7 & 30.2 \\
\ours{} (\emph{ours}) & \textbf{63.9} & \textbf{53.9} & \textbf{58.8} & \textbf{49.1} & \textbf{56.4} \\
\bottomrule
\end{tabular}
\vspace{-5pt}
\caption{
Evaluation results on RGB-D Scenes V2.
\textbf{Boldfaced} numbers highlight the best and the second best are \underline{underlined}.
}
\label{table:results-rgbdv2}
\vspace{-10pt}
\end{table}

\ptitle{Dataset.}
\emph{RGB-D Scenes V2}~\cite{lai2014unsupervised} contains $11427$ RGB-D frames from $14$ indoor scenes. For each scene, we fuse a point cloud fragment with every $25$ consecutive depth frames and sample a RGB image every $25$ frames. We select the image-point-cloud pairs with an overlap ratio of at least $30\%$. The pairs from scenes $1$-$8$ are used for training, scenes $9$ and $10$ for validation, and scenes $11$-$14$ for testing. As last, we obtain a benchmark of $1748$ training pairs, $236$ for validation and $497$ for testing.

\ptitle{Quantative results.}
We first compare our method to the baselines on RGB-D Scenes V2 in \cref{table:results-rgbdv2}. For \emph{Inlier Ratio}, P2-Net outperforms FCGF-2D3D benefiting from the feature saliency-based keypoint detection. However, it still suffers from low inlier ratio.
And albeit achieving better inlier ratio on the first two scenes, Predator-2D3D performs worse in the later two scenes where the camera is closer to the scene.
On the contrary, thanks to the coarse-to-fine matching pipeline and the multi-scale patch pyramid, our \ours{} significantly improves the inlier ratio by $20$ percentage points (pp). And this advantage further contributes to much higher \emph{Feature Matching Recall}, where our method surpasses the second best P2-Net by over $24$ pp.

For the most important \emph{Registration Recall}, P2-Net achieves the best results among the three detection-based baselines. And our method outperforms P2-Net by $18$ pp on registration recall thanks to the more accurate correspondences. These results have demonstrated the strong generality of our method to unseen scenes.

\subsection{Evaluations on 7Scenes}
\label{sec:exp-7scenes}


\begin{table}[!t]
\scriptsize
\setlength{\tabcolsep}{2pt}
\centering
\begin{tabular}{l|cccccccc}
\toprule
Model & Chess & Fire & Heads & Office & Pumpkin & Kitchen & Stairs & Mean \\
\midrule
Mean depth (m) & 1.78 & 1.55 & 0.80 & 2.03 & 2.25 & 2.13 & 1.84 & 1.77 \\
\midrule
\multicolumn{9}{c}{\emph{Inlier Ratio} $\uparrow$} \\
\midrule
FCGF-2D3D~\cite{choy2019fully} & 34.2 & 32.8 & 14.8 & 26.0 & 23.3 & 22.5 & 6.0 & 22.8 \\
P2-Net~\cite{wang2021p2} & \underline{55.2} & \underline{46.7} & 13.0 & \underline{36.2} & \underline{32.0} & \underline{32.8} & 5.8 & \textbf{31.7} \\
Predator-2D3D~\cite{huang2021predator} & 34.7 & 33.8 & \underline{16.6} & 25.9 & 23.1 & 22.2 & \underline{7.5} & 23.4 \\
\ours{} (\emph{ours}) & \textbf{72.1} & \textbf{66.0} & \textbf{31.3} & \textbf{60.7} & \textbf{50.2} & \textbf{52.5} & \textbf{18.1} & \textbf{50.1} \\
\midrule
\multicolumn{9}{c}{\emph{Feature Matching Recall} $\uparrow$} \\
\midrule
FCGF-2D3D~\cite{choy2019fully} & \underline{99.7} & 98.2 & 69.9 & 97.1 & 83.0 & \underline{87.7} & 16.2 & 78.8 \\
P2-Net~\cite{wang2021p2} & \textbf{100.0} & \underline{99.3} & 58.9 & \underline{99.1} & \underline{87.2} & 92.2 & 16.2 & \underline{79.0} \\
Predator-2D3D~\cite{huang2021predator} & 91.3 & 95.1 & \underline{76.7} & 88.6 & 79.2 & 80.6 & \underline{31.1} & 77.5 \\
\ours{} (\emph{ours}) & \textbf{100.0} & \textbf{99.6} & \textbf{98.6} & \textbf{100.0} & \textbf{92.4} & \textbf{95.9} & \textbf{58.1} & \textbf{92.1} \\
\midrule
\multicolumn{9}{c}{\emph{Registration Recall} $\uparrow$} \\
\midrule
FCGF-2D3D~\cite{choy2019fully} & \underline{89.5} & 79.7 & 19.2 & 85.9 & 69.4 & 79.0 & 6.8 & 61.4 \\
P2-Net~\cite{wang2021p2} & \textbf{96.9} & \underline{86.5} & \underline{20.5} & \underline{91.7} & \underline{75.3} & \underline{85.2} & 4.1 & \underline{65.7} \\
Predator-2D3D~\cite{huang2021predator} & 69.6 & 60.7 & 17.8 & 62.9 & 56.2 & 62.6 & \underline{9.5} & 48.5 \\
\ours{} (\emph{ours}) & \textbf{96.9} & \textbf{90.7} & \textbf{52.1} & \textbf{95.5} & \textbf{80.9} & \textbf{86.1} & \textbf{28.4} & \textbf{75.8} \\
\bottomrule
\end{tabular}
\vspace{-5pt}
\caption{
Evaluation results on 7Scenes.
\textbf{Boldfaced} numbers highlight the best and the second best are \underline{underlined}.
}
\label{table:results-7scenes}
\vspace{-10pt}
\end{table}


\begin{figure*}[t]
\centering
\includegraphics[width=\linewidth]{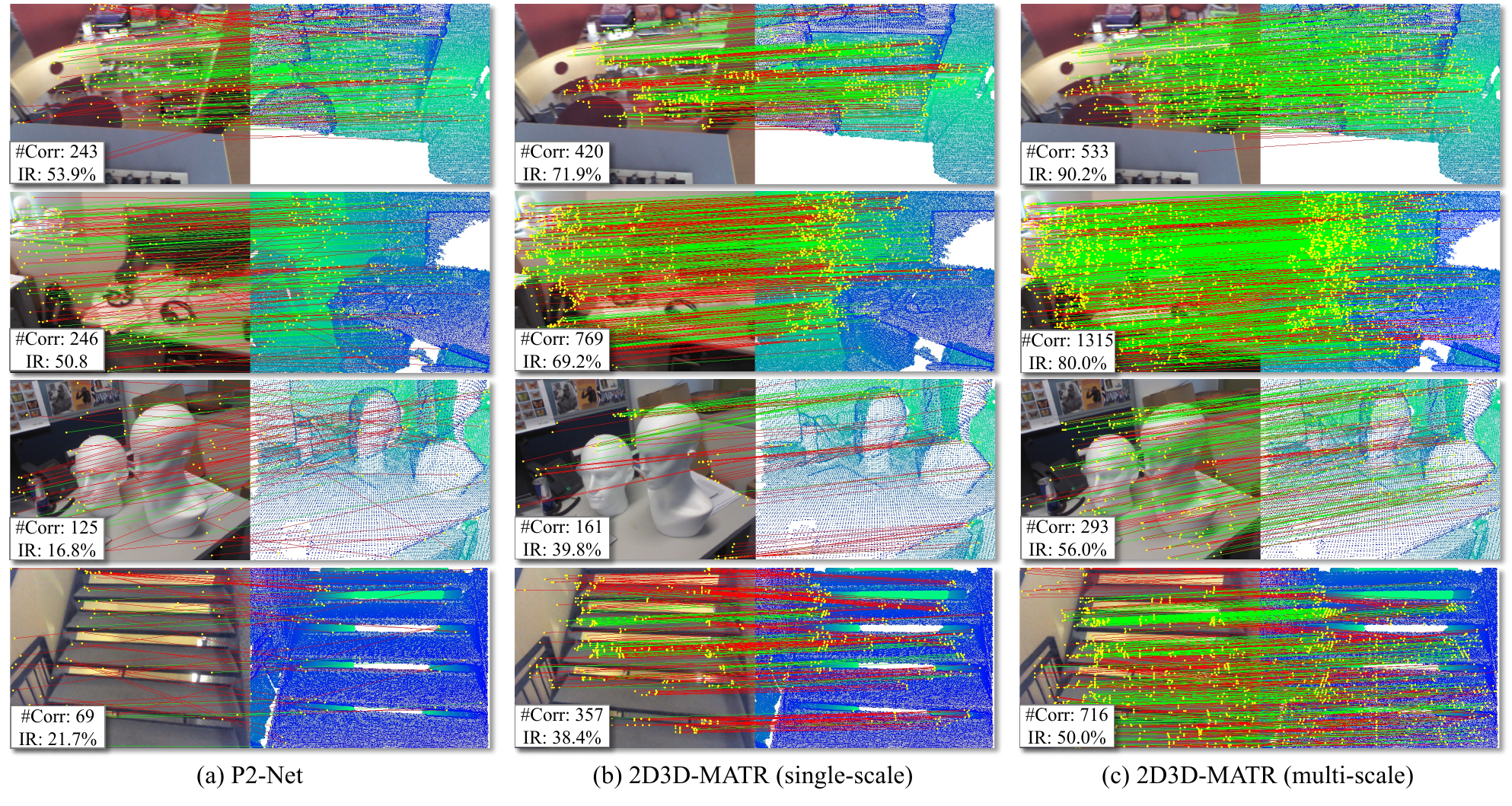}
\vspace{-20pt}
\caption{
Comparisons of correspondences on 7-Scenes. Our method extracts more accurate and more thoroughly distributed correspondences over the whole scene. And it extracts accurate correspondences from repeated patterns (see the $4^{\text{th}}$ row).
}
\label{fig:gallery}
\vspace{-10pt}
\end{figure*}

\ptitle{Dataset.}
\emph{7-Scenes}~\cite{glocker2013real} consists of $46$ RGB-D sequences from $7$ indoor scenes. We use the same method as above to prepare the image-point-cloud pairs from each scene and preserve the pairs that share at least $50\%$ overlap. We follow the official sequence split to generate the training, validation and testing data, which makes $4048$ training pairs, $1011$ validation pairs and $2304$ testing pairs. Note that compared to the benchmark used in~\cite{wang2021p2}, we provide a more challenging one with richer viewpoint changes and smaller overlap ratios. For the evaluation results under the setting of~\cite{wang2021p2}, please refer to Appx.~D.

\ptitle{Quantative results.}
In contrast with~\cref{sec:exp-rgbd}, we evaluate the generality to unseen viewpoints in known scenes on 7-Scenes. The results are demonstrated in~\cref{table:results-7scenes}. For \emph{Inlier Ratio}, our method outperforms the second best P2-Net by over $18$ pp.
For \emph{Feature Matching Recall}, \ours{} achieves an average improvement of $13.1$ pp. And our method surpasses the baseline methods by at least $10$ pp on \emph{Registration Recall}. 
More surprisingly, Predator-2D3D performs the worst on 7-Scenes. As the image-point-cloud pairs in 7-Scenes commonly share more overlap, we assume that explicitly predicting the overlap scores contributes to little benefit but harms the distinctiveness of the learned feature representations.

Compared to RGB-D Scenes V2, 7-Scenes have more significant scale variations across different scenes. Nevertheles, our method still outperforms the baseline methods by a large margin, demonstrating the strong robustness of \ours{} to scale variance.
It is noteworthy that \ours{} achieves more significant improvements on the two hard scenes, \ie, \emph{heads} and \emph{stairs}. On the one hand, the camera is much closer to the scene surfaces in \emph{heads} than in other scenes. This causes great difficulty to extract accurate correspondences as a small error in 3D space could be amplified on the image plane.
On the other hand, \emph{stairs} contains numerous repeated patterns which is hard to distinguish. Thanks to our multi-scale patch pyramid and coarse-to-fine matching strategy, our method can better handle these hard cases.


\begin{table}[!t]
\scriptsize
\centering
\begin{tabular}{l|P{2em}P{2em}P{2em}P{2em}}
\toprule
Model & PIR & IR & FMR & RR \\
\midrule
(a.1) \ours{} (\emph{full}) & 48.5 & \textbf{32.4} & \textbf{90.8} & \textbf{56.4} \\
(a.2) \ours{} w/o coarse-to-fine & - & 11.2 & 52.2 & 34.6 \\
\midrule
(a.1) \ours{} (\emph{full}) & \underline{48.5} & \textbf{32.4} & \underline{90.8} & \textbf{56.4} \\
(b.2) \ours{} w/o self-attention & 45.9 & 29.0 & \textbf{91.8} & 44.0 \\
(b.3) \ours{} w/o cross-attention & \textbf{50.4} & \underline{29.3} & 89.1 & \underline{47.7} \\
(b.4) \ours{} w/o attention & 37.0 & 23.1 & 87.0 & 42.3 \\
\midrule
(c.1) \ours{} (\emph{full}) & \underline{48.5} & \textbf{32.4} & \underline{90.8} & \textbf{56.4} \\
(c.2) \ours{} w/ ($24 \times 32$) & 37.7 & 29.2 & 88.3 & 36.9 \\
(c.3) \ours{} w/ ($12 \times 16$) & 44.2 & 29.9 & 89.2 & 51.7 \\
(c.4) \ours{} w/ ($6 \times 8$) & 41.7 & 23.6 & 87.7 & 50.2 \\
(c.5) \ours{} w/ ($24 \times 32$, $12 \times 16$) & 46.1 & \underline{32.2} & \underline{90.5} & \underline{54.5} \\
(c.6) \ours{} w/ ($24 \times 32$, $6 \times 8$) & 42.3 & 31.6 & 90.0 & 51.3 \\
(c.7) \ours{} w/ ($12 \times 16$, $6 \times 8$) & \textbf{49.8} & 30.9 & 90.1 & 54.2 \\
\bottomrule
\end{tabular}
\vspace{-5pt}
\caption{
Ablation studies on RGB-D Scenes V2.
\textbf{Boldfaced} numbers highlight the best and the second best are \underline{underlined}.
}
\label{table:results-ablation}
\vspace{-10pt}
\end{table}

\ptitle{Qualitative results.}
\cref{fig:gallery} visualizes the extracted correspondences from P2-Net and 2D3D-MATR. We also show the single-scale version of 2D3D-MATR where $24 \times 32$ image patches are used. Our method extracts more accurate and more thoroughly distributed correspondences over the whole scene, which is crucial for successful registration.
The last two rows shows two difficult cases from \emph{heads} and \emph{stairs}.
In the $3^{\text{rd}}$ row, P2-Net fails to detect reliable keypoints and thus suffers from low inlier ratio. Due to a near placement of the camera, the single-scale version of 2D3D-MATR can only extract the correspondences in the distant background areas. On the contrary, benefiting from multi-scale patch pyramid, full 2D3D-MATR extracts much more accurate correspondences distributed over the whole scene.
And the $4^{\text{th}}$ row contains repeated patterns distributed from near to far. P2-Net detects keypoints near the boundaries but fails to match them correctly. Benefiting from the global contextual contraints and cross-modality correlations learned from the transformer module, 2D3D-MATR extracts more accurate correspondences from the stairs. Please refer to Appx.~D for more visualizations.

\subsection{Ablation Studies}
\label{sec:exp-ablation}

We further conduct extensive ablation studies to investigate the efficacy of our designs on RGB-D Scenes V2. Following~\cite{qin2022geometric}, we report another metric \emph{Patch Inlier Ratio} (PIR), the ratio of patch correspondences whose overlap ratios are above a certain threshold (\ie, $0.3$), to evaluate the performance on the coarse level.

\ptitle{Coarse-to-fine matching.}
First, we ablate the coarse-level matching step in our pipeline and match randomly sampled keypoints from both sides as correspondences. In this model, we apply the attention-based feature refinement module between the encoders and the decoders. As shown in \cref{table:results-ablation}(a), the performance drops significantly without the coarse-to-fine matching pipeline. Compared to strict pixel-point matching, patch matching is more robust and reliable as more context could be leveraged. This effectively reduces the searching space during matching, and facilitates extracting accurate correspondences.

\ptitle{Feature refinement module.}
Next, we study the influence of the attention-based feature refinement in \cref{table:results-ablation}(b). We first remove the self-attention modules and the cross-attention modules in the network. The model without self-attention suffers from more serious performance degradation, which means global context plays a more important role than cross-modal aggregation in 2D-3D registration. We then completely remove all attention modules, which further degradates the performance. 

\ptitle{Multi-scale patch pyramid.}
At last, we evaluate the efficicay of the multi-scale patch pyramid in \cref{table:results-ablation}(c). We progressively ablate each resolution level from our full model and evaluate the performance. Obviously, the models with one single resolution perform worse than the multi-scale models, demonstrating the effectiveness of our design. And note that the inlier ratios of the models with small resolution are lower. This is because the image patches in these models are larger and thus leads to more matching ambiguity.


\section{Conclusion}
\label{sec:conclusions}

We have presented \ours{} to hierarchically extract pixel-point correspondences for inter-modality registration between images and point clouds. Benefiting from a coarse-to-fine matching pipeline, our method bypasses the need of keypoint detection across two modalities. We further construct a multi-scale patch pyramid to alleviate the scale ambiguity during patch matching. These designs significantly improve the quality of the extracted correspondences and contribute to accurate 2D-3D registration. A potential limitation of our method is that it still relies on RANSAC for successful registration. In the future, we would like to extend our method for RANSAC-free inter-modality registration.

\ptitle{Acknowledgement.} This work is supported in part by the National Key R\&D Program of China (2018AAA0102200) and the NSFC (62325221, 62132021).


\appendix

\section{Implementation Details}

We mainly compare to three baseline methods in the experiments: (1) FCGF-2D3D, a 2D-3D implementation of FCGF~\cite{choy2019fully}; (2) P2-Net~\cite{wang2021p2}, a 2D-3D version of D2-Net~\cite{dusmanu2019d2} and D3Feat~\cite{bai2020d3feat}; (3) Predator-2D3D, a 2D-3D version of Predator~\cite{huang2021predator}.
For FCGF-2D3D, we supervise the descriptors using circle loss~\cite{sun2020circle} instead of the hardest-in-batch contrastive loss used in~\cite{choy2019fully}. This model could be regarded as a simplified P2-Net without the detection branch. For P2-Net, as there is no official code released for P2-Net, we reimplement it from the scratch. We use the detection loss defined in~\cite{bai2020d3feat} to supervise the detection scores because we find the model fails to converge on our benchmarks using the original detection loss in~\cite{wang2021p2}. For Predator-2D3D, we find that it cannot predict reliable saliency scores in 2D-3D matching, so we only predict the overlapping scores and use them as probabilities to sample random keypoints. And we use transformer~\cite{vaswani2017attention} instead of the graph network in~\cite{huang2021predator} as we find transformer achieves better results. For the baseline methods, we sample $10000$ 2D keypoints and $1000$ 3D keypoints and extract correspondences between them using mutual nearest selection.

\begin{figure*}[t]
\centering
\includegraphics[width=\linewidth]{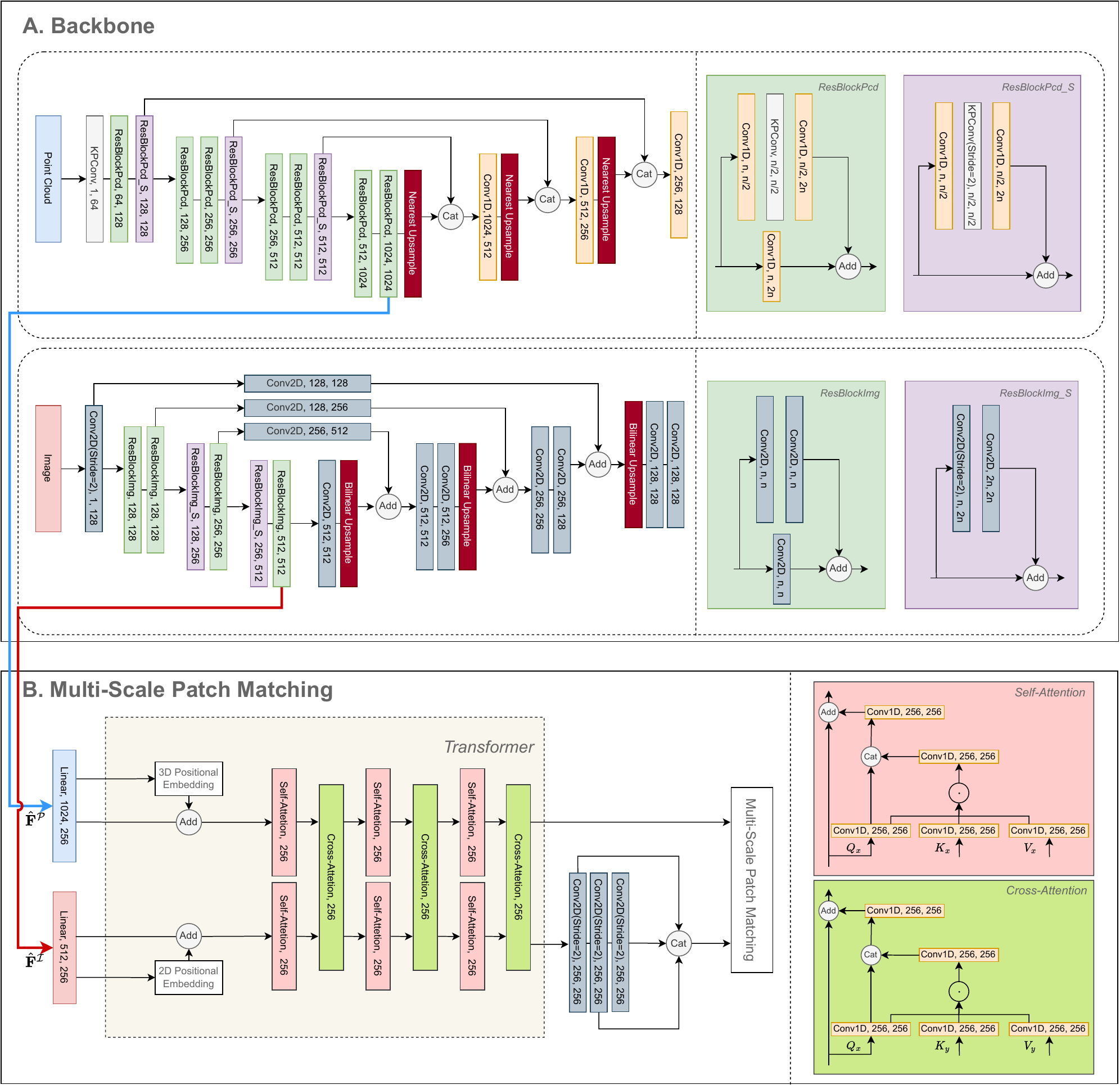}
\caption{
Network architecture.
}
\label{fig:architecture}
\end{figure*}

For fair comparison, we apply the same backbone networks in all the methods, \ie, a $4$-stage ResNet~\cite{he2016deep} with FPN~\cite{lin2017feature} backbone for images and a $4$-stage KPFCNN~\cite{thomas2019kpconv} backbone for point clouds. For the 2D backbone, the output channels of each stage are $\{128, 128, 256, 512\}$. For the 3D backbone, the output channels of each stage are $\{ 128, 256, 512, 1024 \}$. The resolution of the input images is $480 \times 640$ and the resolution in the coarest level is $60 \times 80$. Following~\cite{sun2021loftr}, we convert RGB images to \emph{grayscale} before feeding them to the network. The point clouds are voxelized with an initial voxel size of $2.5$cm and downsampled in each stage using grid subsampling as in~\cite{thomas2019kpconv}. The detailed architecture of our method is illustrated in \cref{fig:architecture}.
And we use the same training settings in all the methods. We use Adam~\cite{kingma2014adam} optimizer to train the networks. The networks are trained for $20$ epochs and the batch size is $1$. The initial learning rate is $10^{-4}$, which is decayed by $0.05$ every epoch.
For all methods (including ours), $256$ correspondences are randomly sampled to supervise the pixel (point) descriptors.
To estimate the transformation, we use PnP-RANSAC implemented in OpenCV~\cite{bradski2000the} with $5000$ iterations and the distance tolerance of $8.0$.


\section{Metrics}

Following~\cite{wang2021p2}, we mainly evaluate our method using $3$ metrics: Inlier Ratio, Feature Matching Recall and Registration Recall.

\emph{Inlier Ratio} (IR) measures the fraction of inliers among all putative pixel-point correspondences.
Following~\cite{wang2021p2}, a correspondence is an inlier if their \emph{3D distance} is below $\tau_1 = 5\text{cm}$ under the ground-truth transformation $\textbf{T}^{*}_{\textbf{P} \rightarrow \textbf{I}}$
\begin{equation}
\mathrm{IR} = \frac{1}{\lvert \mathcal{C} \rvert} \sum_{(\textbf{x}_{i}, \textbf{y}_{i}) \in \mathcal{C}} \llbracket \lVert \textbf{T}^{*}_{\textbf{P} \rightarrow \textbf{I}}(\textbf{x}_{i}) - \mathcal{K}^{-1}(\textbf{y}_{i}) \rVert_2 < \tau_1 \rrbracket,
\end{equation}
where $\llbracket \cdot \rrbracket$ is the Iversion bracket, $\mathbf{x}_{i} \in \mathbf{P}$, $\mathbf{y}_{i} \in \mathbf{Q}$~($\mathbf{Q}$ is the pixel coordinate matrix of $\mathbf{I}$), and $\mathcal{K}^{-1}$ is the function that unprojects a pixel to a 3D point according to its depth value.

\emph{Feature Matching Recall} (FMR) is the fraction of image-point-cloud pairs whose IR is above $\tau_2 = 0.1$.
FMR measures the potential success during the registration:
\begin{equation}
\mathrm{FMR} = \frac{1}{M} \sum_{i=1}^{M} \llbracket \mathrm{IR}_i > \tau_2 \rrbracket,
\end{equation}
where $M$ is the number of all image-point-cloud pairs.

\emph{Registration Recall} (RR) is the fraction of correctly registered testing pairs.
A pair of image and point cloud is regarded as correctly registered if the root mean square error (RMSE) between the point clouds transformed by the ground-truth and the predicted transformation $\textbf{T}_{\textbf{P} \rightarrow \textbf{I}}$ is below $\tau_3 = 0.1\text{m}$:
\begin{align}
\mathrm{RMSE} = & \sqrt{\frac{1}{\lvert \mathbf{P} \rvert} \sum_{\textbf{p}_i \in \mathbf{P}} \lVert \textbf{T}_{\textbf{P} \rightarrow \textbf{I}}(\textbf{p}_{i}) - \textbf{T}^{*}_{\textbf{P} \rightarrow \textbf{I}}(\textbf{p}_{i}) \rVert_2^2}, \\
\mathrm{RR} = & \frac{1}{M} \sum_{i=1}^{M} \llbracket \mathrm{RMSE}_i < \tau_3 \rrbracket.
\end{align}

We further report \emph{Patch Inlier Ratio} (PIR) in the ablation studies to evaluate the accuracy of the patch matching following~\cite{qin2022geometric}.
PIR is the fraction of patch correspondences whose overlap ratios under the ground-truth transformation are above $0.3$.
It reflects the quality of the putative patch correspondences.
A pixel (point) is overlapped if there exists a point (pixel) such that their 3D distance is below a 3D threshold (\ie, $3.75$cm) and their 2D distance is below a 2D threshold (\ie, $8$ pixels). As a result, we obtain two overlap ratios, one on the image side and one on the point cloud side. Here we take the smaller one of them as the final overlap ratio between $\mathbf{I}$ and $\mathbf{P}$.

\section{Data Preparation}

As there is no off-the-shelf benchmarks for 2D-3D registration, we first build two challenging benchmarks based on RGB-D Scenes V2~\cite{lai2014unsupervised} and 7Scenes~\cite{glocker2013real} datasets.

\subsection{RGB-D Scenes V2}
\label{sec:data-rgbd}

RGB-D Scenes V2 consists of RGB-D scans of $14$ indoor scenes. We evaluate the generality to \emph{unseen scenes} of our method and the baselines on this benchmark.
For each scene, we fuse every $25$ consecutive depth frames into a point cloud fragment, which is then voxelized with a voxel size of $2.5$cm. The first RGB image of every $25$ frames are sampled as the set of images.
We then consider every pair of image and point cloud, and select those whose overlap ratios are at least $30\%$. The overlap is computed in the 3D space. The image are first unprojected into a point cloud according to the corresponding depth frame. Then a point is considered as overlapped if there exists a point in the other side which is closer than $3.75$cm to it.
The pairs from scenes $1$-$8$ are used for training, scenes $9$ and $10$ for validation, and scenes $11$-$14$ for testing. As last, we obtain a benchmark of $1748$ training pairs, $236$ for validation and $497$ for testing.
\cref{table:dataset-rgbdv2} shows the statistics on the testing set of our benchmark. In Scene-11 and Scene-12, the camera is further from the scene and the images have a larger range of depth. While in Scene-13 and Scene-14, the scene is much closer to the camera.


\begin{table}[!t]
\scriptsize
\setlength{\tabcolsep}{4pt}
\centering
\begin{tabular}{l|ccccc}
\toprule
Scene & Scene-11 & Scene-12 & Scene-13 & Scene-14 & Mean \\
\midrule
Depth mean (m) & 1.74 & 1.66 & 1.18 & 1.39 & 1.49 \\
Depth std (m) & 0.67 & 0.64 & 0.39 & 0.48 & 0.55 \\
Depth range (m) & 2.20 & 2.22 & 1.72 & 2.07 & 2.05 \\
\bottomrule
\end{tabular}
\vspace{-5pt}
\caption{
Statistics on the testing set of RGB-D Scenes V2.
}
\label{table:dataset-rgbdv2}
\end{table}

\subsection{7-Scenes}
\label{sec:data-7scenes}


\begin{table}[!t]
\scriptsize
\setlength{\tabcolsep}{2.5pt}
\centering
\begin{tabular}{l|cccccccc}
\toprule
Scene & Chess & Fire & Heads & Office & Pumpkin & Kitchen & Stairs & Mean \\
\midrule
Depth mean (m) & 1.78 & 1.55 & 0.80 & 2.03 & 2.25 & 2.13 & 1.84 & 1.77 \\
Depth std (m) & 0.48 & 0.30 & 0.21 & 0.43 & 0.39 & 0.62 & 0.48 & 0.41 \\
Depth range (m) & 2.66 & 1.60 & 0.97 & 1.91 & 1.79 & 2.48 & 3.03 & 2.06 \\
\bottomrule
\end{tabular}
\vspace{-5pt}
\caption{
Statistics on the testing set of 7-Scenes.
}
\label{table:dataset-7scenes}
\end{table}

7-Scenes consists of RGB-D scans of $7$ indoor scenes where each scene has multiple RGB-D sequences. We follow the data split in~\cite{glocker2013real,brachmann2017dsac,wang2021p2} to evaluate the generality to \emph{unseen viewpoints} of our method and the baselines on this benchmark.
For each squence, we follow the same method as in~\cref{sec:data-rgbd} to prepare the point cloud fragments and the RGB image frames. Then, for each scene, we collect the all images and point cloud fragments in the training (testing) sequences, and select the image-point-cloud pairs from them whose overlap ratios are at least $50\%$ as the training (testing) data. The training data are split by $80\%$/$20\%$ for training/validation.
Note that as the RGB images and the depth images are not calibrated in 7-Scenes, we follow~\cite{yang2019sanet} to rescale the image by $\frac{585}{525}$ for an approximate calibration.
\cref{table:dataset-7scenes} shows the statistics on the testing set of 7-Scenes. The distance between the camera and the scene significantly varies in different scenes. The camera is relatively far from the scene in \emph{office}, \emph{pumpkin} and \emph{kitchen}, but is much closer in \emph{heads}. As a result, the scale ambiguity is more significant in 7-Scenes.

\section{Additional Experiments}

\subsection{Additional Ablation Studies}

\ptitle{Patch pyramid.}
In~\cref{table:ablation-rgbdv2-supp}, we further progressively ablate the patch pyramid and report the detailed results on each scene. Note that here all the models are both trained and tested with the corresponding resolution levels, while we albate each pyramid level only in the inference in Tab.~3 of the main paper.

For \emph{Inlier Ratio}, three models achieves comparable results on the first two scenes, but the models with multi-scale patch pyramid performs considerably better than the single-scale one on Scene-13 and Scene-14. As discussed in~\cref{table:dataset-rgbdv2}, the camera is closer to the scene in Scene-13 and Scene-14, which could cause severe inconsistency between the image patchs and the point patches. By leveraging the patch pyramid, the scale ambiguity is alleviated such that more accurate correspondences are obtained.

For \emph{Registration Recall}, more significant improvements are also obtained in the last two scenes. Note that although the three models achieve similar inlier ratios in Scene-11, the multi-scale patch pyramid provide more thoroughly-distributed correpondences, which contributes more accurate registration.


\begin{table}[!t]
\scriptsize
\setlength{\tabcolsep}{3pt}
\centering
\begin{tabular}{l|ccccc}
\toprule
Model & Scene-11 & Scene-12 & Scene-13 & Scene-14 & Mean \\
\midrule
\multicolumn{6}{c}{\emph{Inlier Ratio} $\uparrow$} \\
\midrule
($24 \times 32$, $12 \times 16$, $6 \times 8$) & \textbf{32.8} & \textbf{34.4} & \textbf{39.2} & \textbf{23.3} & \textbf{32.4} \\
($24 \times 32$, $12 \times 16$) & 32.9 & 34.4 & 35.3 & 21.6 & 31.1 \\
($24 \times 32$) & 31.7 & 33.3 & 27.3 & 16.8 & 27.3 \\
\midrule
\multicolumn{6}{c}{\emph{Feature Matching Recall} $\uparrow$} \\
\midrule
($24 \times 32$, $12 \times 16$, $6 \times 8$) & \textbf{98.6} & \textbf{98.0} & \textbf{88.7} & \textbf{77.9} & \textbf{90.8} \\
($24 \times 32$, $12 \times 16$) & 97.2 & 98.0 & 86.6 & 77.0 & 89.7 \\
($24 \times 32$) & 97.2 & 97.1 & 85.6 & 75.7 & 88.9 \\
\midrule
\multicolumn{6}{c}{\emph{Registration Recall} $\uparrow$} \\
\midrule
($24 \times 32$, $12 \times 16$, $6 \times 8$) & \textbf{63.9} & \textbf{53.9} & \textbf{58.8} & \textbf{49.1} & \textbf{56.4} \\
($24 \times 32$, $12 \times 16$) & 55.6 & 53.9 & 43.3 & 41.2 & 48.5 \\
($24 \times 32$) & 52.8 & 51.0 & 26.8 & 26.1 & 39.2 \\
\bottomrule
\end{tabular}
\vspace{-5pt}
\caption{
Additional ablation studies on RGB-D Scenes V2.
\textbf{Boldfaced} numbers highlight the best and the second best are \underline{underlined}.
}
\label{table:ablation-rgbdv2-supp}
\end{table}

\ptitle{Mutual top-$k$ selection.}
We replace the mutual top-$k$ selection in the point matching module with the non-mutual version on RGB-D Scenes V2, which achieves $31.7\%$ IR ($0.7$ pp$\downarrow$), $91.6\%$ FMR ($0.8$ pp$\uparrow$) and $50.8\%$ RR ($5.6$ pp$\downarrow$). We also note that the model with non-mutual top-$k$ selection still beats all the baselines, demonstrating the effectiveness of our method.


\begin{table}[!t]
\scriptsize
\setlength{\tabcolsep}{2pt}
\centering
\begin{tabular}{l|cccccccc}
\toprule
Model & Chess & Fire & Heads & Office & Pumpkin & Kitchen & Stairs & Mean \\
\midrule
\multicolumn{9}{c}{\emph{Inlier Ratio} $\uparrow$} \\
\midrule
FCGF-2D3D~\cite{choy2019fully} & 59.2 & 58.5 & 67.5 & 54.4 & 45.0 & 51.6 & 33.5 & 52.8 \\
P2-Net~\cite{wang2021p2} & 60.9 & 66.9 & 66.1 & 55.8 & \underline{57.0} & 56.1 & 42.4 & 57.9 \\
Predator-2D3D~\cite{huang2021predator} & \underline{75.3} & \underline{71.6} & \textbf{82.1} & \underline{56.1} & 55.3 & \underline{57.2} & \underline{57.7} & \underline{65.0} \\
\ours{} (\emph{ours}) & \textbf{84.1} & \textbf{79.2} & \underline{76.5} & \textbf{73.6} & \textbf{71.8} & \textbf{78.0} & \textbf{69.1} & \textbf{76.0} \\
\midrule
\multicolumn{9}{c}{\emph{Feature Matching Recall} $\uparrow$} \\
\midrule
FCGF-2D3D~\cite{choy2019fully} & 81.8 & 81.0 & 91.0 & 67.5 & 41.7 & 52.3 & 10.5 & 60.8 \\
P2-Net~\cite{wang2021p2} & 82.5 & 93.0 & 89.5 & \underline{70.6} & \underline{76.2} & \underline{64.6} & 22.5 & 71.3 \\
Predator-2D3D~\cite{huang2021predator} & \underline{98.8} & \underline{94.0} & \textbf{100.0} & 66.5 & 69.0 & 61.5 & \underline{69.0} & \underline{79.8} \\
\ours{} (\emph{ours}) & \textbf{100.0} & \textbf{96.5} & \underline{99.0} & \textbf{99.0} & \textbf{92.0} & \textbf{99.5} & \textbf{99.0} & \textbf{97.9} \\
\midrule
\multicolumn{9}{c}{\emph{Registration Recall} $\uparrow$} \\
\midrule
FCGF-2D3D~\cite{choy2019fully} & \underline{99.8} & \textbf{98.0} & 98.0 & 97.0 & 89.2 & 96.7 & \underline{94.5} & 96.2 \\
P2-Net~\cite{wang2021p2} & \underline{99.8} & \textbf{98.0} & 96.0 & \underline{98.1} & \underline{91.7} & \underline{97.2} & 93.0 & \underline{96.3} \\
Predator-2D3D~\cite{huang2021predator} & 99.6 & 92.5 & \textbf{99.0} & 96.5 & 82.0 & 95.5 & 87.0 & 93.2 \\
\ours{} (\emph{ours}) & \textbf{100.0} & \textbf{98.0} & \underline{98.5} & \textbf{98.5} & \textbf{95.0} & \textbf{100.0} & \textbf{98.0} & \textbf{98.3} \\
\bottomrule
\end{tabular}
\vspace{-5pt}
\caption{
Evaluation results on 7Scenes following the experiment settings in~\cite{wang2021p2}.
\textbf{Boldfaced} numbers highlight the best and the second best are \underline{underlined}.
}
\label{table:results-7scenes-easy}
\end{table}

\subsection{Additional Evaluations on 7-Scenes}

We further present the evaluation results on 7-Scenes~\cite{glocker2013real} following the settings in~\cite{wang2021p2}. We fuse a point cloud fragment with $5$ consecutive depth frames. During training, we construct $5$ training pairs between the fused point cloud and the corresponding RGB images. During testing, we only use the last RGB frame to construct $1$ testing pair for each point cloud fragment. The RGB images are rescaled as described in~\cref{sec:data-7scenes}. As a result, we obtain $23500$ training pairs, $2500$ validation pairs, and $3400$ testing pairs. All the models are trained from scratch in the experiments. Compared to our benchmark in the main paper, this setting is more easier due to small transformation and high overlap ratio. Note that the thresholds for the metrics in this setting are $\tau_1 = 4.5\text{cm}$, $\tau_2 = 50\%$ and $\tau_3 = 5\text{cm}$ following~\cite{wang2021p2}.

The results are shown in~\cref{table:results-7scenes-easy}. For \emph{Inlier Ratio}, \ours{} outperforms the baseline methods by a large margin, especially on the last four harder scenes. This further contributes to significant improvements on \emph{Feature Matching Recall}, where our method surpasses the second best Predator-2D3D by $18$ pp. For \emph{Registration Recall}, the performance tends to be saturated in most scenes. Nonetheless, \ours{} still achieves the best results, especially on \emph{pumpkin} and \emph{stairs}. These results have demonstrated the efficacy of our method.


\begin{figure*}[t]
\centering
\includegraphics[width=0.9\linewidth]{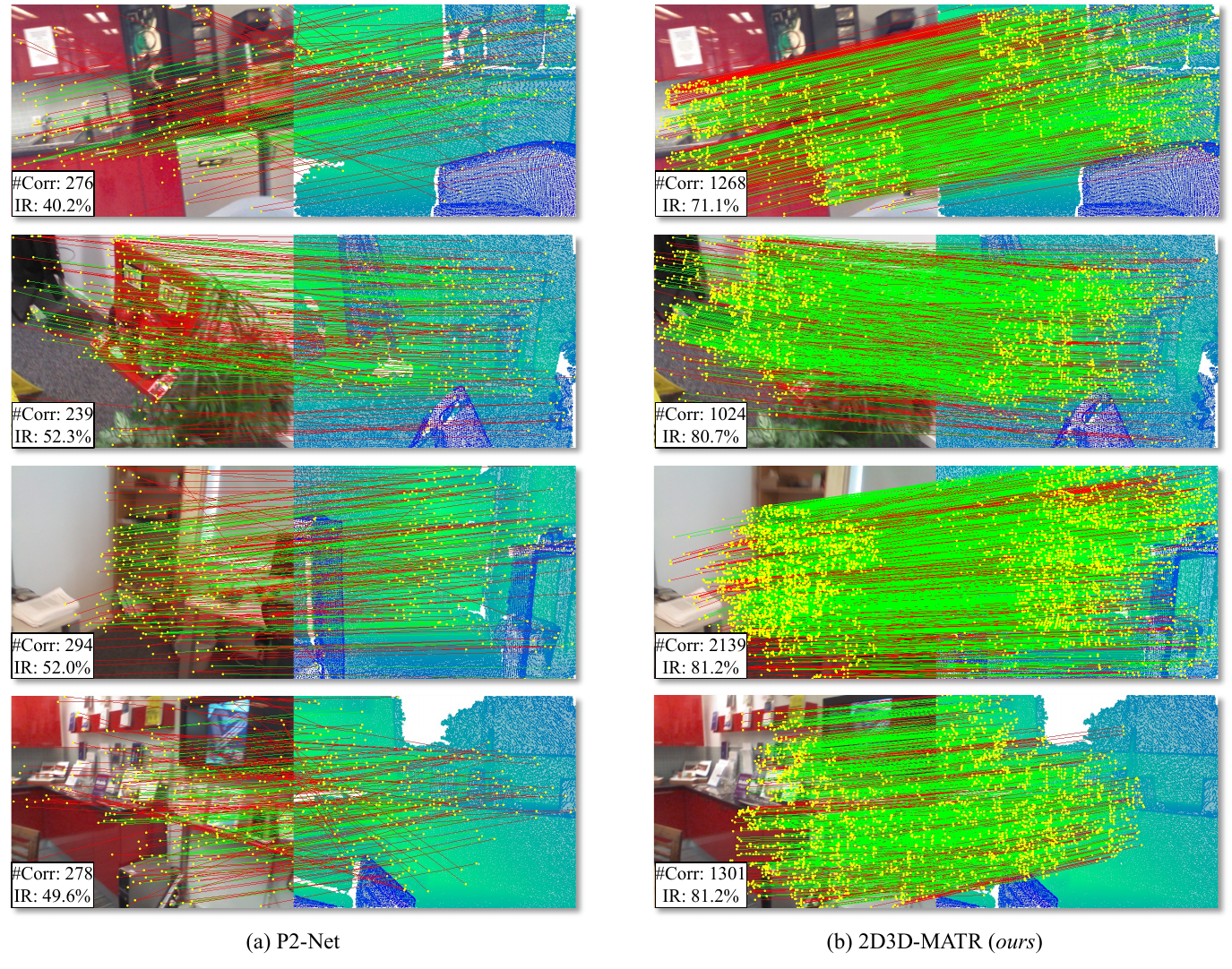}
\caption{
Comparisons of extracted correspondences on 7-Scenes.
}
\label{fig:7scenes-gallery-supp}
\end{figure*}


\begin{figure*}[t]
\centering
\includegraphics[width=0.9\linewidth]{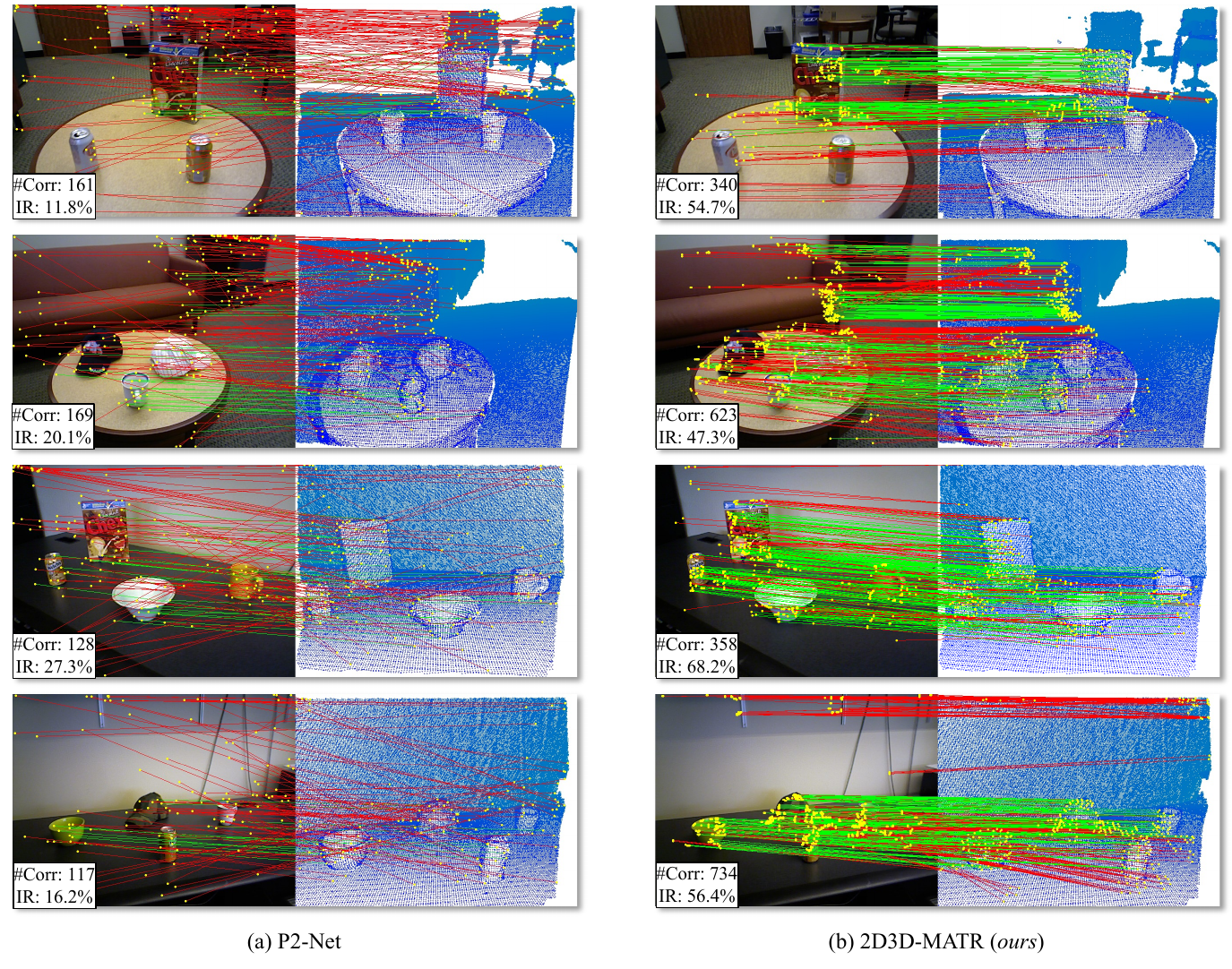}
\caption{
Comparisons of extracted correspondences on RGB-D Scenes V2.
}
\label{fig:rgbd-gallery-supp}
\end{figure*}

\subsection{Qualitative Results}

We provide more qualitative comparisons of P2-Net~\cite{wang2021p2} and \ours{} on 7-Scenes (\cref{fig:7scenes-gallery-supp}) and RGB-D Scenes V2 (\cref{fig:rgbd-gallery-supp}). It is observed that the correspondences from our method are much denser and more accurate those from P2-Net. Moreover, \ours{} extracts correspondences from both near and far regions, showing strong robustness to scale variance.

\section{Limitations}

Albeit achieving the new state-of-the-art preformance, \ours{} could have the following three limitations.

First, despite of higher inlier ratio, our method still rely on PnP-RANSAC to estimate the alignment transformation. Compared to point cloud registration, the 2D errors of the 2D-3D correspondences are sensitive to the camera pose. For instance, given two points which are $5$cm away from each other in the 3D space, their distance in the image plane could be merely $2$ pixels if they are far from the camera but up to tens of pixels if they are close to the camera. For this reason, it is more difficult to achieve accurate registration and thus PnP-RANSAC is still necessary.

Second, we find that the generality of 2D-3D matching to novel scenes is not as good as that of image matching or point cloud matching. This can be observed by comparing the results on RGB-D Scenes V2 and 7-Scenes, where the former is worse. We assume the reason is that inter-modality matching is more difficult than intra-modality matching as one need project the visual information from different domains to a common feature space.

Third, the uniform patch partition strategy in our method is relatively simple and coarse. Although we design a multi-scale patch pyramid mechanism to handle scale ambiguity, the patches are still not perfectly aligned in most cases. This could cause difficulty in learning consistent features for the patches, and increase redundancy in the fine-level matching.
A possible solution is to leverage semantic information to extract patches, which we will leave as future work.

{\small
\bibliographystyle{ieee_fullname}
\bibliography{2d3dmatr}

\begin{thebibliography}{10}\itemsep=-1pt

\bibitem{bai2021pointdsc}
Xuyang Bai, Zixin Luo, Lei Zhou, Hongkai Chen, Lei Li, Zeyu Hu, Hongbo Fu, and
  Chiew-Lan Tai.
\newblock Pointdsc: Robust point cloud registration using deep spatial
  consistency.
\newblock In {\em CVPR}, pages 15859--15869, 2021.

\bibitem{bai2020d3feat}
Xuyang Bai, Zixin Luo, Lei Zhou, Hongbo Fu, Long Quan, and Chiew-Lan Tai.
\newblock D3feat: Joint learning of dense detection and description of 3d local
  features.
\newblock In {\em CVPR}, pages 6359--6367, 2020.

\bibitem{brachmann2017dsac}
Eric Brachmann, Alexander Krull, Sebastian Nowozin, Jamie Shotton, Frank
  Michel, Stefan Gumhold, and Carsten Rother.
\newblock Dsac-differentiable ransac for camera localization.
\newblock In {\em CVPR}, pages 6684--6692, 2017.

\bibitem{brachmann2016uncertainty}
Eric Brachmann, Frank Michel, Alexander Krull, Michael~Ying Yang, Stefan
  Gumhold, et~al.
\newblock Uncertainty-driven 6d pose estimation of objects and scenes from a
  single rgb image.
\newblock In {\em CVPR}, pages 3364--3372, 2016.

\bibitem{brachmann2018learning}
Eric Brachmann and Carsten Rother.
\newblock Learning less is more-6d camera localization via 3d surface
  regression.
\newblock In {\em CVPR}, pages 4654--4662, 2018.

\bibitem{brachmann2019neural}
Eric Brachmann and Carsten Rother.
\newblock Neural-guided ransac: Learning where to sample model hypotheses.
\newblock In {\em CVPR}, pages 4322--4331, 2019.

\bibitem{bradski2000the}
G. Bradski.
\newblock The opencv library.
\newblock {\em Dr. Dobb's Journal of Software Tools}, 2000.

\bibitem{chopra2005learning}
Sumit Chopra, Raia Hadsell, and Yann LeCun.
\newblock Learning a similarity metric discriminatively, with application to
  face verification.
\newblock In {\em CVPR}, volume~1, pages 539--546. IEEE, 2005.

\bibitem{choy2020deep}
Christopher Choy, Wei Dong, and Vladlen Koltun.
\newblock Deep global registration.
\newblock In {\em CVPR}, pages 2514--2523, 2020.

\bibitem{choy2019fully}
Christopher Choy, Jaesik Park, and Vladlen Koltun.
\newblock Fully convolutional geometric features.
\newblock In {\em ICCV}, pages 8958--8966, 2019.

\bibitem{deng2018ppf}
Haowen Deng, Tolga Birdal, and Slobodan Ilic.
\newblock Ppf-foldnet: Unsupervised learning of rotation invariant 3d local
  descriptors.
\newblock In {\em ECCV}, pages 602--618, 2018.

\bibitem{deng2018ppfnet}
Haowen Deng, Tolga Birdal, and Slobodan Ilic.
\newblock Ppfnet: Global context aware local features for robust 3d point
  matching.
\newblock In {\em CVPR}, pages 195--205, 2018.

\bibitem{detone2018superpoint}
Daniel DeTone, Tomasz Malisiewicz, and Andrew Rabinovich.
\newblock Superpoint: Self-supervised interest point detection and description.
\newblock In {\em CVPRW}, pages 224--236, 2018.

\bibitem{drost2010model}
Bertram Drost, Markus Ulrich, Nassir Navab, and Slobodan Ilic.
\newblock Model globally, match locally: Efficient and robust 3d object
  recognition.
\newblock In {\em CVPR}, pages 998--1005. Ieee, 2010.

\bibitem{dusmanu2019d2}
Mihai Dusmanu, Ignacio Rocco, Tomas Pajdla, Marc Pollefeys, Josef Sivic,
  Akihiko Torii, and Torsten Sattler.
\newblock D2-net: A trainable cnn for joint description and detection of local
  features.
\newblock In {\em CVPR}, pages 8092--8101, 2019.

\bibitem{feng20192d3d}
Mengdan Feng, Sixing Hu, Marcelo~H Ang, and Gim~Hee Lee.
\newblock 2d3d-matchnet: Learning to match keypoints across 2d image and 3d
  point cloud.
\newblock In {\em ICRA}, pages 4790--4796. IEEE, 2019.

\bibitem{fischler1981random}
Martin~A Fischler and Robert~C Bolles.
\newblock Random sample consensus: a paradigm for model fitting with
  applications to image analysis and automated cartography.
\newblock {\em Communications of the ACM}, 24(6):381--395, 1981.

\bibitem{glocker2013real}
Ben Glocker, Shahram Izadi, Jamie Shotton, and Antonio Criminisi.
\newblock Real-time rgb-d camera relocalization.
\newblock In {\em ISMAR}, pages 173--179. IEEE, 2013.

\bibitem{gojcic2019perfect}
Zan Gojcic, Caifa Zhou, Jan~D Wegner, and Andreas Wieser.
\newblock The perfect match: 3d point cloud matching with smoothed densities.
\newblock In {\em CVPR}, pages 5545--5554, 2019.

\bibitem{he2016deep}
Kaiming He, Xiangyu Zhang, Shaoqing Ren, and Jian Sun.
\newblock Deep residual learning for image recognition.
\newblock In {\em CVPR}, pages 770--778, 2016.

\bibitem{hoffer2014deep}
Elad Hoffer and Nir Ailon.
\newblock Deep metric learning using triplet network.
\newblock {\em arXiv preprint arXiv:1412.6622}, 2014.

\bibitem{huang2021predator}
Shengyu Huang, Zan Gojcic, Mikhail Usvyatsov, Andreas Wieser, and Konrad
  Schindler.
\newblock Predator: Registration of 3d point clouds with low overlap.
\newblock In {\em CVPR}, pages 4267--4276, 2021.

\bibitem{kingma2014adam}
Diederik~P Kingma and Jimmy Ba.
\newblock Adam: A method for stochastic optimization.
\newblock In {\em ICLR}, 2015.

\bibitem{lai2014unsupervised}
Kevin Lai, Liefeng Bo, and Dieter Fox.
\newblock Unsupervised feature learning for 3d scene labeling.
\newblock In {\em ICRA}, pages 3050--3057. IEEE, 2014.

\bibitem{lee2021deep}
Junha Lee, Seungwook Kim, Minsu Cho, and Jaesik Park.
\newblock Deep hough voting for robust global registration.
\newblock In {\em ICCV}, pages 15994--16003, 2021.

\bibitem{lee2021patchmatch}
Jae~Yong Lee, Joseph DeGol, Victor Fragoso, and Sudipta~N Sinha.
\newblock Patchmatch-based neighborhood consensus for semantic correspondence.
\newblock In {\em CVPR}, pages 13153--13163, 2021.

\bibitem{lepetit2009epnp}
Vincent Lepetit, Francesc Moreno-Noguer, and Pascal Fua.
\newblock Epnp: An accurate o (n) solution to the pnp problem.
\newblock {\em IJCV}, 81(2):155--166, 2009.

\bibitem{li2018so}
Jiaxin Li, Ben~M Chen, and Gim~Hee Lee.
\newblock So-net: Self-organizing network for point cloud analysis.
\newblock In {\em CVPR}, pages 9397--9406, 2018.

\bibitem{li2020dual}
Xinghui Li, Kai Han, Shuda Li, and Victor Prisacariu.
\newblock Dual-resolution correspondence networks.
\newblock {\em NeurIPS}, 33:17346--17357, 2020.

\bibitem{li2020hierarchical}
Xiaotian Li, Shuzhe Wang, Yi Zhao, Jakob Verbeek, and Juho Kannala.
\newblock Hierarchical scene coordinate classification and regression for
  visual localization.
\newblock In {\em CVPR}, pages 11983--11992, 2020.

\bibitem{li2018full}
Xiaotian Li, Juha Ylioinas, and Juho Kannala.
\newblock Full-frame scene coordinate regression for image-based localization.
\newblock {\em arXiv preprint arXiv:1802.03237}, 2018.

\bibitem{lin2017feature}
Tsung-Yi Lin, Piotr Doll{\'a}r, Ross Girshick, Kaiming He, Bharath Hariharan,
  and Serge Belongie.
\newblock Feature pyramid networks for object detection.
\newblock In {\em CVPR}, pages 2117--2125, 2017.

\bibitem{lowe1999object}
David~G Lowe.
\newblock Object recognition from local scale-invariant features.
\newblock In {\em ICCV}, volume~2, pages 1150--1157. Ieee, 1999.

\bibitem{luo2020aslfeat}
Zixin Luo, Lei Zhou, Xuyang Bai, Hongkai Chen, Jiahui Zhang, Yao Yao, Shiwei
  Li, Tian Fang, and Long Quan.
\newblock Aslfeat: Learning local features of accurate shape and localization.
\newblock In {\em CVPR}, pages 6589--6598, 2020.

\bibitem{massiceti2017random}
Daniela Massiceti, Alexander Krull, Eric Brachmann, Carsten Rother, and
  Philip~HS Torr.
\newblock Random forests versus neural networks—what's best for camera
  localization?
\newblock In {\em ICRA}, pages 5118--5125. IEEE, 2017.

\bibitem{meng2017backtracking}
Lili Meng, Jianhui Chen, Frederick Tung, James~J Little, Julien Valentin, and
  Clarence~W de Silva.
\newblock Backtracking regression forests for accurate camera relocalization.
\newblock In {\em IROS}, pages 6886--6893. IEEE, 2017.

\bibitem{meng2018exploiting}
Lili Meng, Frederick Tung, James~J Little, Julien Valentin, and Clarence~W de
  Silva.
\newblock Exploiting points and lines in regression forests for rgb-d camera
  relocalization.
\newblock In {\em IROS}, pages 6827--6834. IEEE, 2018.

\bibitem{mildenhall2020nerf}
Ben Mildenhall, Pratul~P Srinivasan, Matthew Tancik, Jonathan~T Barron, Ravi
  Ramamoorthi, and Ren Ng.
\newblock Nerf: Representing scenes as neural radiance fields for view
  synthesis.
\newblock In {\em ECCV}, pages 405--421, 2020.

\bibitem{pham2020lcd}
Quang-Hieu Pham, Mikaela~Angelina Uy, Binh-Son Hua, Duc~Thanh Nguyen, Gemma
  Roig, and Sai-Kit Yeung.
\newblock Lcd: Learned cross-domain descriptors for 2d-3d matching.
\newblock In {\em AAAI}, volume~34, pages 11856--11864, 2020.

\bibitem{qin2022geometric}
Zheng Qin, Hao Yu, Changjian Wang, Yulan Guo, Yuxing Peng, and Kai Xu.
\newblock Geometric transformer for fast and robust point cloud registration.
\newblock In {\em CVPR}, pages 11143--11152, 2022.

\bibitem{revaud2019r2d2}
Jerome Revaud, Philippe Weinzaepfel, C{\'e}sar~De Souza, and Martin
  Humenberger.
\newblock R2d2: repeatable and reliable detector and descriptor.
\newblock In {\em NeurIPS}, pages 12414--12424, 2019.

\bibitem{rocco2020efficient}
Ignacio Rocco, Relja Arandjelovi{\'c}, and Josef Sivic.
\newblock Efficient neighbourhood consensus networks via submanifold sparse
  convolutions.
\newblock In {\em ECCV}, pages 605--621. Springer, 2020.

\bibitem{rocco2018neighbourhood}
Ignacio Rocco, Mircea Cimpoi, Relja Arandjelovi{\'c}, Akihiko Torii, Tomas
  Pajdla, and Josef Sivic.
\newblock Neighbourhood consensus networks.
\newblock {\em NeurIPS}, 31, 2018.

\bibitem{rublee2011orb}
Ethan Rublee, Vincent Rabaud, Kurt Konolige, and Gary Bradski.
\newblock Orb: An efficient alternative to sift or surf.
\newblock In {\em ICCV}, pages 2564--2571. Ieee, 2011.

\bibitem{rusu2009fast}
Radu~Bogdan Rusu, Nico Blodow, and Michael Beetz.
\newblock Fast point feature histograms (fpfh) for 3d registration.
\newblock In {\em ICRA}, pages 3212--3217. IEEE, 2009.

\bibitem{sarlin2020superglue}
Paul-Edouard Sarlin, Daniel DeTone, Tomasz Malisiewicz, and Andrew Rabinovich.
\newblock Superglue: Learning feature matching with graph neural networks.
\newblock In {\em CVPR}, pages 4938--4947, 2020.

\bibitem{shotton2013scene}
Jamie Shotton, Ben Glocker, Christopher Zach, Shahram Izadi, Antonio Criminisi,
  and Andrew Fitzgibbon.
\newblock Scene coordinate regression forests for camera relocalization in
  rgb-d images.
\newblock In {\em CVPR}, pages 2930--2937, 2013.

\bibitem{sun2021loftr}
Jiaming Sun, Zehong Shen, Yuang Wang, Hujun Bao, and Xiaowei Zhou.
\newblock Loftr: Detector-free local feature matching with transformers.
\newblock In {\em CVPR}, pages 8922--8931, 2021.

\bibitem{sun2020circle}
Yifan Sun, Changmao Cheng, Yuhan Zhang, Chi Zhang, Liang Zheng, Zhongdao Wang,
  and Yichen Wei.
\newblock Circle loss: A unified perspective of pair similarity optimization.
\newblock In {\em CVPR}, pages 6398--6407, 2020.

\bibitem{thomas2019kpconv}
Hugues Thomas, Charles~R Qi, Jean-Emmanuel Deschaud, Beatriz Marcotegui,
  Fran{\c{c}}ois Goulette, and Leonidas~J Guibas.
\newblock Kpconv: Flexible and deformable convolution for point clouds.
\newblock In {\em ICCV}, pages 6411--6420, 2019.

\bibitem{valentin2015exploiting}
Julien Valentin, Matthias Nie{\ss}ner, Jamie Shotton, Andrew Fitzgibbon,
  Shahram Izadi, and Philip~HS Torr.
\newblock Exploiting uncertainty in regression forests for accurate camera
  relocalization.
\newblock In {\em CVPR}, pages 4400--4408, 2015.

\bibitem{vaswani2017attention}
Ashish Vaswani, Noam Shazeer, Niki Parmar, Jakob Uszkoreit, Llion Jones,
  Aidan~N Gomez, {\L}ukasz Kaiser, and Illia Polosukhin.
\newblock Attention is all you need.
\newblock {\em NeurIPS}, 30, 2017.

\bibitem{wang2021p2}
Bing Wang, Changhao Chen, Zhaopeng Cui, Jie Qin, Chris~Xiaoxuan Lu, Zhengdi Yu,
  Peijun Zhao, Zhen Dong, Fan Zhu, Niki Trigoni, et~al.
\newblock P2-net: Joint description and detection of local features for pixel
  and point matching.
\newblock In {\em ICCV}, pages 16004--16013, 2021.

\bibitem{yang2019sanet}
Luwei Yang, Ziqian Bai, Chengzhou Tang, Honghua Li, Yasutaka Furukawa, and Ping
  Tan.
\newblock Sanet: Scene agnostic network for camera localization.
\newblock In {\em ICCV}, pages 42--51, 2019.

\bibitem{yu2021cofinet}
Hao Yu, Fu Li, Mahdi Saleh, Benjamin Busam, and Slobodan Ilic.
\newblock Cofinet: Reliable coarse-to-fine correspondences for robust
  pointcloud registration.
\newblock {\em NeurIPS}, 34:23872--23884, 2021.

\bibitem{zhou2021patch2pix}
Qunjie Zhou, Torsten Sattler, and Laura Leal-Taixe.
\newblock Patch2pix: Epipolar-guided pixel-level correspondences.
\newblock In {\em CVPR}, pages 4669--4678, 2021.

\end{thebibliography}
}

\end{document}